\PassOptionsToPackage{dvipsnames}{xcolor}
\documentclass{article}

\usepackage{amsfonts}
\usepackage{amsmath}
\usepackage{bbm}
\usepackage{comment}
\usepackage{microtype}
\usepackage{multirow}
\usepackage{graphicx}
\usepackage{subcaption}
\usepackage{booktabs} 
\usepackage{tabularx}
\usepackage{amsthm}
\usepackage{hhline}
\usepackage[inline]{enumitem}

\usepackage{hyperref}

\DeclareMathOperator*{\argmin}{arg\,min}

\newcommand\tp[2][-6]{{#2}^{\mkern#1mu\top}} 

\setlist[enumerate,1]{label={\arabic*)}}

\usepackage[accepted]{icml2024}

\usepackage{amsmath}
\usepackage{amssymb}
\usepackage{mathtools}
\usepackage{amsthm}

\usepackage[capitalize,noabbrev]{cleveref}

\theoremstyle{plain}

\theoremstyle{definition}

\theoremstyle{remark}

\usepackage[textsize=tiny]{todonotes}

\icmltitlerunning{Layerwise Proximal Replay}

\begin{document}

\twocolumn[
\icmltitle{Layerwise Proximal Replay: A Proximal Point Method\\for Online Continual Learning}



\icmlsetsymbol{equal}{*}

\begin{icmlauthorlist}
\icmlauthor{Jason Yoo}{ubc}
\icmlauthor{Yunpeng Liu}{ubc,inverted}
\icmlauthor{Frank Wood}{ubc,inverted,mila}
\icmlauthor{Geoff Pleiss}{ubc,vector}
\end{icmlauthorlist}

\icmlaffiliation{ubc}{University of British Columbia}
\icmlaffiliation{inverted}{Inverted AI}
\icmlaffiliation{mila}{Mila}
\icmlaffiliation{vector}{Vector Institute}

\icmlcorrespondingauthor{Jason Yoo}{jasony97@cs.ubc.ca}

\icmlkeywords{Machine Learning, ICML}

\vskip 0.3in
]



\printAffiliationsAndNotice{}  

\begin{abstract}
In online continual learning, a neural network incrementally learns from a non-i.i.d. data stream.
Nearly all online continual learning methods employ experience replay to simultaneously prevent catastrophic forgetting and underfitting on past data.
Our work demonstrates a limitation of this approach:
neural networks trained with experience replay tend to have unstable optimization trajectories, impeding their overall accuracy.
Surprisingly, these instabilities persist even when the replay buffer stores all previous training examples,
suggesting that this issue is orthogonal to catastrophic forgetting.
We minimize these instabilities through a simple modification of the optimization geometry.
Our solution, Layerwise Proximal Replay (LPR), balances learning from new and replay data while only allowing for gradual changes in the hidden activation of past data.
We demonstrate that LPR consistently improves replay-based online continual learning methods across multiple problem settings, regardless of the amount of available replay memory.
\end{abstract}


\section{Introduction}

Continual learning is a subfield of machine learning that studies how to enable models to continuously adapt to new information over time without forgetting old information under memory and computation constraints. Online continual learning is a particularly challenging form of continual learning where a neural network must learn from a  non-stationary data stream.
Data arrive in small batches, typically between $1$ to $10$ data points, and the network can typically only train for a few iterations before the next batch arrives.

Almost all successful approaches to online continual learning rely on \emph{experience replay} \citep{chaudhry2019continual},
which retains a small subset of prior data used during training to approximate the loss over all past data \citep{soutifcormerais2023comprehensive}.
Replay-based approaches are particularly dominant in online continual learning for two reasons.
First, training on prior data helps to prevent catastrophic forgetting \citep{french1999catastrophic}, arguably the primary bottleneck in continual learning.
Second, replay increases the overall number of training iterations where the network is exposed to past data.
This is especially crucial in the online setting, which is otherwise prone to underfitting due to the rapid arrival of new data and in many cases the inability to train models to convergence \citep{zhang2022simple}.

Despite their proven ability to mitigate catastrophic forgetting and underfitting,
we hypothesize that online replay methods are inefficient from an optimization perspective.
At any given training iteration,
the new data batch and replay data both contribute to the overall training loss.
However, since the network has already trained to some extent on replay data,
new data will in general contribute more to the overall training loss.
The resulting parameter update will then bias the network toward the new data and cause the network function to forget its behavior on prior data, at times severely \citep{caccia2021new,de2022continual}.
Although past data performance can be partially recovered after additional training on replay examples, this optimization-time predictive instability can adversely affect optimization efficiency and overall model accuracy.

In this work, we tackle these inefficiencies through a simple optimizer modification,
which maintains the advantages of replay-based continual learning while improving its stability.
Our method, which we refer to as \emph{Layerwise Proximal Replay} ({\bf LPR}),
applies a replay-tailored preconditioner to the loss gradients and
builds upon a growing literature of continual learning methods that modify the gradient update geometry \citep{chaudhry2018efficient,zeng2019continual,kao2021natural,saha2021gradient,saha2023continual}.
LPR's preconditioner balances two desiderata:
\begin{enumerate*}
    \item maximally learning from \emph{both new and past data} while
    \item only gradually changing past data predictions (and internal representations) to promote optimization stability and efficiency.
\end{enumerate*}

Across several predictive tasks and datasets,
we find that LPR consistently improves upon the accuracy of replay-based online continual learning methods.
Importantly, LPR achieves performance gains that are robust to the choice of replay loss as well as the size of the replay buffer.
We even notice substantial performance gains when the replay buffer has unlimited memory and thus is able to ``remember'' all data encountered during training.
These results suggest that LPR can improve performance even when the catastrophic forgetting problem is largely mitigated.

To summarize, we make the following contributions:
\begin{enumerate}
    \item We introduce Layerwise Proximal Replay (LPR), the first online continual learning method that combines experience replay with a proximal point method.
    \item We demonstrate through extensive experimentation that LPR consistently improves a wide range of state-of-the-art online continual learning methods across problem settings and replay buffer sizes.
    \item We analyze the effects of LPR, demonstrating improved optimization and fewer destabilizing changes to the network's internal representations.
\end{enumerate}

\section{Background}

\paragraph{Notation.}
Given an $L$-layer neural network,
we will denote a specific layer's parameters as the matrix $\Theta^{(\ell)}$, where $\ell$ is the layer index.
For linear layers, $\Theta^{(\ell)}$ is a $d^{\ell} \times d^{\ell + 1}$ matrix, where $d^{\ell}$ is the dimensionality of layer $\ell$'s hidden representation.
The vector $\theta$ will refer to the concatenation of all of the (flattened) network parameters---i.e. $\theta = \begin{bmatrix} \mathrm{vec}(\Theta^{(1)}) & \ldots & \mathrm{vec}(\Theta^{(L)}) \end{bmatrix}$.

\paragraph{Online continual learning} is a problem setup where a model incrementally learns from a potentially non-i.i.d. data stream in a memory and/or computation limited setting. In the classification setting, at time $\tau$ the neural network receives a data batch $\mathcal{D}_{\tau} = \{X_{\tau,i}, Y_{\tau,i}\}_{i=1}^{n_{\tau}}$ with $n_{\tau}$ features and labels. We assume that we are in the classification setting and that $n_{\tau}$ is equal for all data batches for the rest of the paper. Once the neural network trains on a data batch, it cannot train on the same batch again aside from a small sample of it that may be present in a replay buffer.

In the online continual learning literature, the data stream is often built from a sequence of ``tasks'', where each task is defined by a dataset a model should train on. The data stream is built as follows. Initially, each task's dataset is split into mutually exclusive data batches of much smaller size that are i.i.d. sampled without replacement. Then, a data stream is constructed such that it sequentially returns all data batches from the first task, then all data batches from the second task, and so forth.

Online continual learning is distinct from offline continual learning in a number of ways \citep{soutifcormerais2023comprehensive}. Typically, $n_{\tau}$ is much smaller than offline continual learning, falling in between 1 and 100. In addition, if the data stream is characterized by a sequence of tasks (a problem accompanied by a dataset) where data batches that belong to a single tasks arrive in an i.i.d. fashion, the task boundary is usually not provided.
Lastly, models must perform well at any point of their training, unlike offline continual learning models where evaluation is performed every time when the models finish training on a task. 

In practice, models underfit the data stream in online continual learning \citep{zhang2022simple} because they can only train on each data batch's datapoints \textit{once}, aside from a small number of past datapoints that may be kept in a replay buffer. To clarify, the model may take multiple contiguous gradient steps for the new data batch, but it may no longer be trained on the same data batch afterward.


\paragraph{Replay methods} are employed in both offline and online continual learning, but they are especially dominant in the latter \citep{soutifcormerais2023comprehensive}. At time $\tau$, replay methods loosely approximate the loss over all previously observed data batches with
\begin{align}
    \mathcal{L}(\theta) \coloneqq \mathcal{L}_\mathrm{new}(\theta, \mathcal{D}_{\tau}) + \alpha \mathcal{L}_{\mathrm{replay}}(\theta, \mathcal{M}) \label{eq:er_obj}
\end{align}
where $\theta$ is the vector of network parameters, $\mathcal{M}$ is either a much smaller subset of past data batches or pseudo datapoints that summarize the past data, $\alpha$ is the replay regularization strength, and $\mathcal{L}_{\mathrm{replay}}(\theta, \mathcal{M})$ is the loss associated with the replay buffer. We note that the forms of $\mathcal{L}_\mathrm{new}$ and $\mathcal{L}_\mathrm{replay}$ do not have to be identical.

\paragraph{Proximal point methods} \citep{drusvyatskiy2017proximal,parikh2019proximal,censor1992proximal} are optimization algorithms that minimize a function $\mathcal{L}(\theta)$ by repeated application of the $\mathrm{prox}$ operator:
\begin{equation}
    \begin{aligned}
        \theta_{j+1} &= \mathrm{prox}_{\eta \mathcal L} \left( \theta_j \right),
        \\
        \mathrm{prox}_{\eta \mathcal L} \left( \theta_j \right) &:=
        \argmin_{\theta} \left[ \mathcal{L}(\theta) + \tfrac{1}{2\eta} \Vert \theta - \theta_j \Vert^2 \right]. \label{eq:prox_general}
    \end{aligned}
\end{equation}
Here,  $\theta_{j}$ is the model parameter value at $j$-th optimization iteration, $\Vert \cdot \Vert$ is the proximal penalty norm\footnote{One can also use a Bregman divergence instead of a norm \citep{censor1992proximal}.}, and $\eta^{-1}$ is the proximal penalty strength.
For many functions, solving Equation \ref{eq:prox_general} can be as hard as solving the original minimization objective. Therefore, it is common to make approximations such as linearization of $\mathcal{L}$ around $\theta_j$.
Notably, gradient descent can be cast as a proximal point method by linearizing $\mathcal{L}$ around $\theta_j$ and using the Euclidean norm as the proximal penalty.

\section{Layerwise Proximal Replay}
\label{sec:method}

\paragraph{Motivation.}
We have three desiderata for replay-based online continual learning:
\begin{enumerate*}
    \item the network should rapidly learn from new data batches,
    \item the network should continue learning from prior (replay) data, and
    \item parameter updates should not cause sudden unnecessary performance degradation on the past data, thus ensuring stable optimization while preventing underfitting.
\end{enumerate*}
We hypothesize that existing online replay methods largely ignore the third, resulting in inefficient optimization that ultimately harms accuracy.

We can instead satisfy all three desiderata by modifying the optimizer on top of experience replay.
Consider the standard SGD update $\theta_{j+1} = \theta_j - \eta \nabla \mathcal L(\theta_j)$.
This update naturally codifies the first two desiderata,
as $\mathcal L(\theta_j)$ combines both the new data and replay data losses.
To encode the third, we simply constrain the optimizer to only consider updates that minimally change replay predictions.

Consider the proximal formulation of SGD \citep[e.g.][]{parikh2019proximal}:
\begin{align}
    \theta_{j+1} &= \mathrm{prox}_{\eta \mathcal{L}} \left( \theta_j \right)
    \nonumber
    := \argmin_{\theta} \left( \mathcal L(\theta) + \tfrac{1}{2\eta} \Vert \theta - \theta_j \Vert^2_2 \right)
    \\
    &\approx
        \argmin_{\theta} \left( \left\langle \nabla \mathcal L(\theta_j), \theta \! - \! \theta_j \right\rangle + \tfrac{1}{2\eta} \Vert \theta - \theta_j \Vert^2_2 \right)
    \label{eqn:grad_descent_prox}
    \\
    &= \theta_j - \eta \nabla \mathcal L(\theta_j)
    \nonumber
\end{align}
where \cref{eqn:grad_descent_prox} uses first order Taylor approximation $\mathcal L(\theta) - \mathcal L(\theta_j) \approx \left\langle \nabla \mathcal L(\theta_j), \theta - \theta_j \right\rangle$.
Intuitively, the norm penalty on $\theta - \theta_j$ ensures that the update only considers values of $\theta$ where the Taylor approximation is reasonable.
We can further restrict the search space to updates that minimally change replay predictions by modifying \cref{eqn:grad_descent_prox} as follows:
\begin{equation}
    \begin{aligned}
        &\argmin_{\theta}
        \left( \langle \nabla \mathcal L(\theta_j), \theta - \theta_j \rangle + \tfrac{1}{2\eta} \Vert \theta - \theta_j \Vert^2_2 \right),\\
        &\ \mathrm{s.t.}\ \Vert F(\theta) - F(\theta_j) \Vert_F^2 < \delta.
    \end{aligned}
    \label{eq:constraint}
\end{equation}
Here, $F(\theta) \in \mathbb R^{|\mathcal M| \times c}$ is the matrix of predictions (ex. logits) for replay data given neural network parameters $\theta$, $\Vert \cdot \Vert_F$ is the Frobenius norm, and $\delta > 0$ is a hyperparameter that penalizes changes in replay data predictions.

\paragraph{Layerwise Proximal Replay.}
In practice, it is challenging to implement this functional constraint efficiently in online continual learning.
To that end, we instead consider the stronger constraint
of ensuring a minimal change to the \emph{hidden activations} of replay data:
\begin{equation}
    \begin{aligned}
        &\argmin_{\theta}
        \left( \left\langle \nabla \mathcal L(\theta_j), \theta - \theta_j \right\rangle + \tfrac{1}{2\eta} \Vert \theta - \theta_j \Vert^2_2 \right),\\
        &\ \mathrm{s.t.}\ \Vert Z^{(\ell)}(\theta) - Z^{(\ell)}(\theta_j) \Vert_F^2 < \delta^{(\ell)}, \:\: \ell \in \{1, \ldots, L\}
    \end{aligned}
    \label{eqn:lpr_constrained}
\end{equation}
where $Z^{(\ell)}(\theta) \in \mathbb R^{|\mathcal M| \times d^{\ell}}$
is the matrix of $\ell$-th layer activations for all replay data and
$\delta^{(\ell)} > 0$ are some layerwise constraint constants. We can approximately apply this constraint in a layer-wise fashion.
Let $\Theta^{(i)} \in \mathbb R^{d^{\ell} \times d^{\ell + 1}}$ be the matrix of parameters at layer $\ell$
and let $Z^{(\ell)}_j \in \mathbb R^{|\mathcal M| \times d^{\ell}}$ be the hidden activations for layer $\ell$ at time $j$.
The activations follow the recursive formula $Z^{(\ell+1)}_j = \phi^{(\ell)}( Z^{(\ell)}_j \Theta^{(\ell)}_j )$
where $\phi^{(\ell)}$ is the $l$-th layer's activation function.
If our parameter update minimally changes layer $\ell$'s activations (i.e. $Z^{(\ell)}_{j+1} \approx Z^{(\ell)}_{j}$), we can rewrite Equation \ref{eq:constraint}'s constraint as
\[
   \Vert Z^{(\ell)}_{j} \Theta^{(\ell)} - Z^{(\ell)}_{j} \Theta^{(\ell)}_{j} \Vert_F^2 < \delta^{(\ell)}.
\]
Applying a Lagrange multiplier to transform \cref{eqn:lpr_constrained}, we have
\begin{equation}
    \begin{aligned}
    \argmin_{\theta} \big( &\left\langle \nabla \mathcal L(\theta_j), \theta - \theta_j \right\rangle + \tfrac{1}{2\eta} \Vert \theta - \theta_j \Vert^2_2 \label{eq:lagrange}\\
    &+ {\textstyle \sum_{\ell=1}^L } \lambda^{(\ell)} \Vert Z^{(\ell)}_{j} \Theta^{(\ell)} - Z^{(\ell)}_{j} \Theta^{(\ell)}_{j} \Vert_F^2 \big).
    \end{aligned}
\end{equation}
This can in turn be converted to a layer-wise update rule
\begin{align}
    &\begin{aligned}
    \Theta^{(\ell)}_{j+1} 
    \! = \!
    \argmin_{\Theta^{(\ell)}} \!
    \bigg(& \! \left\langle \nabla \mathcal{L}(\Theta_j^{(\ell)}), \Theta^{(\ell)} \!\! - \!\Theta_j^{(\ell)\!} \right\rangle \! \\
    & + \! \tfrac{1}{2\eta} \Vert \Theta^{(\ell)} \!\! - \!\Theta^{(\ell)}_j\! \Vert^2_{P_\ell} \bigg)
    \end{aligned} \label{eqn:lpr_prox}
    \\
    \! &\qquad= \!
    \Theta_j^{(\ell)} - \eta P_\ell^{-1} \nabla L(\Theta_j^{(\ell)}),
    \label{eqn:lpr}
\end{align}
where $P_\ell$ is a positive definite matrix given by
\begin{equation}
    P_\ell := I + \omega^{\ell} Z^{(\ell)^\top}_j Z^{(\ell)}_j,
    \label{eqn:preconditioner}
\end{equation}
$\omega^{\ell} = 2\eta\lambda^{(\ell)}$ is a hyperparameter that absorbs the Lagrange multiplier $\lambda^{(\ell)}$, $\nabla L(\Theta_j^{(\ell)})$ is a shorthand for $\nabla_{\Theta^{(\ell)}} \mathcal L(\theta_j)$,
and $\Vert \cdot \Vert_{P_\ell}$ is the norm given by $\sqrt{\mathrm{tr} \left( (\cdot)^\top P_\ell (\cdot) \right)}$
(See Appendix \ref{appendix:preconditioner_proof} for the derivation.)

Note that \cref{eqn:lpr_prox}, like \cref{eqn:grad_descent_prox}, can be construed as an application of the $\mathrm{prox}$ operator,
but here the proximity function is given by the layerwise $P_\ell$-norm rather than the standard Euclidean norm.
The use of this non-Euclidean proximal update in conjunction with the replay loss ensures continued learning
while simultaneously limiting sudden degradations in replay performance.
We refer to \cref{eqn:lpr} as \emph{Layerwise Proximal Replay}, or LPR for short.

\begin{algorithm}[tb]
   \caption{Layerwise Proximal Replay (LPR)}
   \label{alg:algorithm}
    \begin{algorithmic}
        \STATE {\bfseries Input:} network parameters $\theta$, learning rate $\eta$, per batch gradient steps count $S$, preconditioner update interval $T$, layerwise stability hyperparameters $\omega_0^{\ell}$ for $\ell \in \{1,...,L\}$.
        \STATE {\bfseries Output:} trained network parameters $\theta$.
        \\\hrulefill
        \STATE Initialize  replay buffer: $\mathcal{M} \leftarrow \{\}$.
        \STATE Initialize preconditioner inverses $\Lambda_{\ell} \leftarrow I,\ \ell \in \{1,...,L\}$. {\color{gray}\COMMENT{$\Lambda_\ell = P_\ell^{-1}$}}
        \FOR{$\tau \in \{1,...,\infty\}$}
            \STATE Obtain new data batch $\mathcal{D}_{\tau}$.
            \FOR{$s \in \{1,...,S\}$}
                \STATE Compute  loss $\mathcal{L}(\theta)$ using $\mathcal{D}_{\tau}$ and $\mathcal{M}$.
                \STATE Compute  loss gradient $\nabla \mathcal{L}(\theta)$.
                \FOR{layer index $\ell \in \{1,...,L\}$}
                    \STATE $\Theta^{(\ell)} \leftarrow \Theta^{(\ell)} - \eta \Lambda_{\ell}\nabla_{\Theta^{(\ell)}} \mathcal{L}(\theta)$ 
                \ENDFOR
            \ENDFOR
            \STATE Update $\mathcal{M}$ with $\mathcal{D}_{\tau}$.
            \IF{$\tau \bmod{T} = 0$}
                \STATE Obtain  feature tensor $X_{mem}$ from $\mathcal{M}$.
                \STATE Set $Z^{(1)}$ to $X_{mem}$ and $n$ to the batch size of $X_{mem}$.
                \FOR{layer index $\ell \in \{1,...,L\}$}
                    \STATE $\omega^{\ell} \leftarrow \omega_0^{\ell} / n$
                    \STATE $\Lambda_{\ell} \leftarrow \left(\omega^{\ell}\tp{Z^{(\ell)}}Z^{(\ell)} + I\right)^{-1}$ 
                    \STATE $Z^{(\ell+1)} \leftarrow \phi^{(\ell)}(Z^{(\ell)}\Theta^{(\ell)})$
                \ENDFOR
            \ENDIF
       \ENDFOR
    \end{algorithmic}
\end{algorithm}

\paragraph{Analysis.}
LPR can be viewed as a variant of preconditioned SGD
where the symmetric positive definite preconditioner $P_\ell$ is specific to each layer.
We note that setting $\omega^{\ell} = 0$ recovers the standard SGD update.
When $\omega^{\ell} > 0$, the eigenvalues of $P_\ell$ are all greater than 1,
and thus $P_\ell^{-1}$ is a contraction operator; i.e.
\[
    \Vert P_\ell^{-1} \nabla \mathcal L(\Theta_j^{(\ell)}) \Vert_F < \Vert \nabla \mathcal{L}(\Theta_j^{(\ell)}) \Vert_F.
\]
\setlength{\textfloatsep}{5.5mm}

Crucially, this contraction is not a simple rescaling of the gradient contributions from new data.
Intuitively, we have $\Vert P_\ell^{-1} \nabla \mathcal L(\Theta_j^{(\ell)}) \Vert_F \ll \Vert \nabla \mathcal{L}(\Theta_j^{(\ell)}) \Vert_F$
when the standard gradient update would significantly alter replay predictions,
and $\Vert P_\ell^{-1} \nabla \mathcal L(\Theta_j^{(\ell)}) \Vert_F < \Vert \nabla \mathcal{L}(\Theta_j^{(\ell)}) \Vert_F$ otherwise.


\paragraph{Implementation.}
Algorithm \ref{alg:algorithm} illustrates LPR in detail.
The hyperparameter $S$ and $T$ specifies how many gradient steps to take per incoming data batch and the interval at which $P_\ell$ is recomputed from $\mathcal{M}$ for computational purposes respectively.
Algorithmically,
we store our preconditioner as $\Lambda_\ell := P_\ell^{-1}$
so that each application of the preconditioner does not require performing matrix inversion.

In practice, specifying the hyperparameters $\omega_0^{\ell}$ for each layer can be cumbersome, especially for large neural networks. Therefore, in our experiments we parameterize $\omega_0^{\ell}$ for all layers using two scalar values $\omega_0, \beta$ (refer to Appendix \ref{appendix:omega_param}). We specify the $\ell$-th layer's proximal penalty strength via $\omega^\ell = \omega_0^\ell/n$, where $n$ is the number of replay buffer data used to compute $P_\ell^{-1}$. This decouples $\omega^\ell$ from replay buffer data size and ensures that the network is plastic enough to continue learning from new data.

In online continual learning, optimal hidden activations for past data are subject to change when additional relevant observations arrive from the data stream. Thus, LPR periodically refreshes $P_{\ell}$ with the current neural network's feed-forward activations of the replay buffer inputs $X_{mem}$ from $\mathcal{M}$.
For neural network layers that are not linear layers but can be cast as linear transformations, for example convolution and batch normalization, the input activation $Z^{(\ell)}$ is reshaped to allow it to be matrix multiplied with the layer's weight matrix $\Theta^{(\ell)}$ for computing the layer's output.

Compared to regular experience replay, LPR requires one additional matrix multiplication per layer for preconditioning the layer's gradient per gradient step. In addition, once every $T$ data batches, it must feed forward replay buffer data through the network and invert $d^{\ell} \times d^{\ell}$ matrices per layer. The only memory requirement on top of regular experience replay is keeping $L$ preconditioners $P_{\ell}$.

\paragraph{Relation to projection-based continual learning.}
Our method adds to a growing body of continual learning methods that apply transformations to the gradient updates \citep{chaudhry2018efficient,zeng2019continual,farajtabar2020orthogonal,saha2021gradient,kao2021natural,deng2021flattening,shu2022replay,guo2022adaptive,saha2023continual}.
We offer a detailed review of these methods in \cref{sec:extended_related_work}.
Most gradient transformation methods have only been considered for offline continual learning (with \citet{chaudhry2018efficient, guo2022adaptive, hess2023two} being notable exceptions).
Despite many surface level similarities, LPR differs from existing transformations in subtle but crucial ways necessary for replay-based online continual learning.

Our work is particularly related to those that construct the projection matrix such that the gradient updates occur approximately orthogonal to the subspace spanned by the past data's hidden activations \citep{zeng2019continual,saha2021gradient,deng2021flattening,lin2022trgp,shu2022replay,zhao2023rethinking,saha2023continual}.
At a high level, these methods try to limit parameter updates to a $(d^{\ell} - k)$-dimensional subspace by applying an orthogonal projection to the gradients:
$(I - \Phi^{(\ell)} (\Phi^{(\ell)^\top} \Phi^{(\ell)})^{-1} \Phi^{(\ell)^\top}) \nabla \mathcal L(\Theta^{(\ell)})$,
where the matrix $\Phi^{(\ell)} \in \mathbb R^{d^{\ell} \times k}$ defines the nullspace of the projection operator.
Some work sets $\Phi^{(\ell)}$ to a low-rank approximation of all past-data activations computed in an SVD-like fashion \citep{saha2021gradient, deng2021flattening, lin2022trgp, saha2023continual} while others average mini-batch activations to slowly build up $\Phi^{\ell}$ and not run out of capacity \citep{zeng2019continual,guo2022adaptive}.
All methods structure the projection matrix such that gradient updates will preserve the high-level statistics of past-data representations as much as possible.

It is worth noting that projection operations (and their close approximations) are largely incompatible with continually learning from replay buffers.
To see this, note that the matrix $(I - \Phi^{(\ell)} (\Phi^{(\ell)^\top} \Phi^{(\ell)})^{-1} \Phi^{(\ell)^\top})$ will project out any gradient contributions from data points spanned by $\Phi^{(\ell)}$. (See \cref{appendix:replay_grad_projections} for a derivation.)
If $\Phi^{(\ell)}$ approximately spans prior activations,
then the replay loss will largely be projected out.
Of course, $\Phi^{(\ell)}$ is too low dimensional to span all prior activations, and most practical methods use a ``soft'' approximation to the projection operator \citep{zeng2019continual,deng2021flattening,saha2023continual}.
Nevertheless, gradient projection methods largely minimize gradient information from the replay buffer, which is undesirable when we want to continue learning on past data.

LPR's preconditioner has subtle but crucial differences that make it more compatible with online replay losses.
Applying the Woodbury inversion formula to \cref{eqn:lpr,eqn:preconditioner}, the LPR update can be written as
\begin{equation}
\left( I - Z^{(\ell)} \left( \tfrac{1}{\omega^{(\ell)}} I + Z^{(\ell)^\top} Z^{(\ell)} \right)^{-1} Z^{(\ell)^\top} \right) \nabla \mathcal L(\Theta^{(\ell)})
\label{eqn:woodbury}
\end{equation}
Note that we only recover a projection operator in the limiting case where
\begin{enumerate*}
    \item the replay buffer is really small (i.e. $\vert \mathcal M \vert < d^{\ell}$, so that $Z^{(\ell)}$ spans a subspace of $\mathbb R^{d^{\ell}}$) and
    \item $\omega^{(\ell)} \to \infty$.
\end{enumerate*}
Neither of these conditions is desirable for replay-based online continual learning.
First, setting $\omega^{(\ell)} \to \infty$ would prevent any changes to replay data activations,
which is undesirable in the online setting since networks are not trained to convergence before data are added to the replay buffer.
We note that LPR with very high values of $\omega^\ell$ were never selected during our hyperparameter optimization (refer to \cref{appendix:experiment}).
Second, limiting $\vert \mathcal M \vert < d^{\ell}$ severely harms performance, as prior work demonstrates that replay methods improve significantly as $\vert \mathcal{M} \vert$ gets larger.
Altogether, LPR promotes further learning on replay data compared to projection methods and does not compress prior activations into low-rank approximations.

\begin{table*}[tbp]
    \centering
    \resizebox{0.95\textwidth}{!}{%
    \begin{tabularx}{0.975\textwidth}{|>{\centering\arraybackslash}X||>{\centering\arraybackslash}X|>{\centering\arraybackslash}X|>{\centering\arraybackslash}X||>{\centering\arraybackslash}X|>{\centering\arraybackslash}X|>{\centering\arraybackslash}X|}
        \hline
        & \multicolumn{3}{c||}{Memory Size 1000} & \multicolumn{3}{c|}{Memory Size 2000}\\
        \hline
        Method & Acc & AAA & WC-Acc & Acc & AAA & WC-Acc\\
        \hline
        ER & 0.2262 {\scriptsize ± 0.0042} & 0.3255 {\scriptsize ± 0.0055} & 0.1072 {\scriptsize ± 0.0022} & 0.2937 {\scriptsize ± 0.0054} & 0.3586 {\scriptsize ± 0.0048} & 0.1264 {\scriptsize ± 0.0030}\\
        LPR (ER) & \textbf{0.2631} {\scriptsize ± 0.0042} & \textbf{0.3782} {\scriptsize ± 0.0059} & \textbf{0.1522} {\scriptsize ± 0.0020} & \textbf{0.3334} {\scriptsize ± 0.0056} & \textbf{0.4251} {\scriptsize ± 0.0047} & \textbf{0.1926} {\scriptsize ± 0.0026}\\
        \hline
        DER & 0.2339 {\scriptsize ± 0.0045} & 0.3333 {\scriptsize ± 0.0059} & 0.1178 {\scriptsize ± 0.0035} & 0.2935 {\scriptsize ± 0.0039} & 0.3575 {\scriptsize ± 0.0073} & 0.1267 {\scriptsize ± 0.0036}\\
        LPR (DER) & \textbf{0.2634} {\scriptsize ± 0.0036} & \textbf{0.3982} {\scriptsize ± 0.0053} & \textbf{0.1700} {\scriptsize ± 0.0029} & \textbf{0.3292} {\scriptsize ± 0.0053} & \textbf{0.4325} {\scriptsize ± 0.0059} & \textbf{0.2018} {\scriptsize ± 0.0032}\\
        \hline
        EMA & 0.3124 {\scriptsize ± 0.0061} & 0.3346 {\scriptsize ± 0.0063} & 0.1078 {\scriptsize ± 0.0029} & 0.3725 {\scriptsize ± 0.0039} & 0.3665 {\scriptsize ± 0.0050} & 0.1234 {\scriptsize ± 0.0030}\\
        LPR (EMA) & \textbf{0.3289} {\scriptsize ± 0.0039} & \textbf{0.3783} {\scriptsize ± 0.0063} & \textbf{0.1454} {\scriptsize ± 0.0024} & \textbf{0.3884} {\scriptsize ± 0.0029} & \textbf{0.4135} {\scriptsize ± 0.0046} & \textbf{0.1755} {\scriptsize ± 0.0032}\\
        \hline
        LODE & 0.2442 {\scriptsize ± 0.0054} & 0.3391 {\scriptsize ± 0.0068} & 0.1150 {\scriptsize ± 0.0033} & 0.2969 {\scriptsize ± 0.0033} & 0.3728 {\scriptsize ± 0.0072} & 0.1346 {\scriptsize ± 0.0036}\\
        LPR (LODE) & \textbf{0.2740} {\scriptsize ± 0.0051} & \textbf{0.3785} {\scriptsize ± 0.0061} & \textbf{0.1753} {\scriptsize ± 0.0054} & \textbf{0.3261} {\scriptsize ± 0.0049} & \textbf{0.4102} {\scriptsize ± 0.0058} & \textbf{0.2014} {\scriptsize ± 0.0045}\\
        \hline
    \end{tabularx}
    }
    \caption{Split-CIFAR100 results with mean and standard error computed across 10 random seeds.}
    \label{tab:c100}
\end{table*}

\begin{table*}[tbp]
    \centering
    \resizebox{0.95\textwidth}{!}{%
    \begin{tabularx}{0.975\textwidth}{|>{\centering\arraybackslash}X||>{\centering\arraybackslash}X|>{\centering\arraybackslash}X|>{\centering\arraybackslash}X||>{\centering\arraybackslash}X|>{\centering\arraybackslash}X|>{\centering\arraybackslash}X|}
        \hline
        & \multicolumn{3}{c||}{Memory Size 2000} & \multicolumn{3}{c|}{Memory Size 4000}\\
        \hline
        Method & Acc & AAA & WC-Acc & Acc & AAA & WC-Acc\\
        \hline
        ER & 0.1590 {\scriptsize ± 0.0016} & 0.2883 {\scriptsize ± 0.0047} & 0.1016 {\scriptsize ± 0.0016} & 0.2192 {\scriptsize ± 0.0021} & 0.3368 {\scriptsize ± 0.0044} & 0.1499 {\scriptsize ± 0.0014}\\
        LPR (ER) & \textbf{0.1669} {\scriptsize ± 0.0014} & \textbf{0.2990} {\scriptsize ± 0.0051} & \textbf{0.1134} {\scriptsize ± 0.0010} & \textbf{0.2312} {\scriptsize ± 0.0023} & \textbf{0.3492} {\scriptsize ± 0.0042} & \textbf{0.1623} {\scriptsize ± 0.0016}\\
        \hline
        DER & 0.1611 {\scriptsize ± 0.0020} & 0.3010 {\scriptsize ± 0.0041} & 0.1206 {\scriptsize ± 0.0010} & 0.2294 {\scriptsize ± 0.0019} & 0.3457 {\scriptsize ± 0.0045} & 0.1685 {\scriptsize ± 0.0016}\\
        LPR (DER) & \textbf{0.1801} {\scriptsize ± 0.0031} & \textbf{0.3121} {\scriptsize ± 0.0047} & \textbf{0.1294} {\scriptsize ± 0.0016} & \textbf{0.2408} {\scriptsize ± 0.0032} & \textbf{0.3530} {\scriptsize ± 0.0051} & \textbf{0.1751} {\scriptsize ± 0.0013}\\
        \hline
        EMA & 0.1900 {\scriptsize ± 0.0022} & 0.2870 {\scriptsize ± 0.0040} & 0.1024 {\scriptsize ± 0.0014} & 0.2603 {\scriptsize ± 0.0017} & 0.3369 {\scriptsize ± 0.0038} & 0.1506 {\scriptsize ± 0.0013}\\
        LPR (EMA) & \textbf{0.1959} {\scriptsize ± 0.0018} & \textbf{0.3017} {\scriptsize ± 0.0039} & \textbf{0.1128} {\scriptsize ± 0.0013} & \textbf{0.2671} {\scriptsize ± 0.0016} & \textbf{0.3504} {\scriptsize ± 0.0039} & \textbf{0.1629} {\scriptsize ± 0.0013}\\
        \hline
        LODE & \textbf{0.1978} {\scriptsize ± 0.0022} & 0.3132 {\scriptsize ± 0.0045} & 0.1351 {\scriptsize ± 0.0015} & 0.2452 {\scriptsize ± 0.0022} & 0.3522 {\scriptsize ± 0.0041} & 0.1731 {\scriptsize ± 0.0015}\\
        LPR (LODE) & 0.1971 {\scriptsize ± 0.0019} & \textbf{0.3170} {\scriptsize ± 0.0044} & \textbf{0.1360} {\scriptsize ± 0.0017} & \textbf{0.2518} {\scriptsize ± 0.0015} & \textbf{0.3592} {\scriptsize ± 0.0049} & \textbf{0.1799} {\scriptsize ± 0.0015}\\
        \hline
    \end{tabularx}
    }
    \caption{Split-TinyImageNet results with mean and standard error computed across 10 random seeds.}
    \label{tab:tin}
\end{table*}

\section{Experiments}

In this section, we empirically investigate LPR's effect on how a neural network learns during online continual learning. In addition, we extensively evaluate LPR across three online continual learning problems on top of four state-of-the-art experience replay methods.\footnote{The code is available at \href{https://github.com/plai-group/LPR}{https://github.com/plai-group/LPR}.}

\subsection{Setup}

\paragraph{Experiments.}
We build on top of the Avalanche continual learning framework \citep{carta2023avalanche} and closely follow the experiment setup from \citet{soutifcormerais2023comprehensive}. We investigate LPR's behavior and accuracy on two online class-incremental learning and one online domain-incremental learning problem settings \citep{van2022three}. For online class-incremental learning, we evaluate on the online versions of Split-CIFAR100 and Split-TinyImageNet datasets \citep{soutifcormerais2023comprehensive}. Each dataset is divided and ordered into a sequence of 20 tasks that have mutually exclusive labels.
These sequences are then converted to data streams where every data batch from the stream contains 10 datapoints that are i.i.d. sampled within tasks but not across tasks.
For online domain-incremental learning, we evaluate on the online version of the CLEAR dataset \citep{lin2021clear}. Domain-incremental learning involves adapting to sequentially changing data domains in real time. CLEAR's built-in sequence of 10 tasks is converted to a data stream in the same manner as the online versions of Split-CIFAR100 and Split-TinyImageNet datasets. The task ordering is dependent on the random seed for all datasets.

\paragraph{Models.}
We use slim and full versions of ResNet18 for online class-incremental and domain-incremental experiments respectively.
Following \citet{soutifcormerais2023comprehensive}, we train all models using SGD with no momentum and no weight decay.
For each data batch, we take 3, 9, and 10 gradient steps respectively for Split-CIFAR100, Split-TinyImageNet, and Online CLEAR. Experience replay methods randomly sample 10 exemplars from the replay buffer and compute the loss on both the current data batch and the replay exemplars per gradient step.
For models using LPR, the preconditioner is updated every 10 data batches, using the entire replay buffer for memory constrained experiments and 2,000-4,000 randomly sampled memory exemplars for memory-unconstrained experiments. Following \citet{zhang2022simple}, we apply image augmentations randomly to exemplars.
We discuss hyperparameter selection and additional experiment details in \cref{appendix:experiment}. We additionally report LPR's training time and sensitivity analysis in \cref{appendix:sensitivity}.

\paragraph{Replay losses.}
To assess the generality of LPR, we apply it on top of four state-of-the-art experience replay methods: experience replay (ER) \citep{chaudhry2019continual}, dark experience replay++ (DER) \citep{buzzega2020dark}, exponential moving average (EMA) \citep{carta2023improving}, and loss decoupling (LODE) \citep{liang2023loss}.

\paragraph{Other baselines.}
We also report the results of two gradient projection-based online continual learning methods A-GEM \citep{chaudhry2019continual} and AOP \citep{guo2022adaptive}. Given their significantly suboptimal performance compared to experience replay, we have included the results for these models across three datasets in Appendix \ref{appendix:gp_results}.

\paragraph{Metrics.}
We measure the test set final accuracy {\bf (Acc)}, the validation set average anytime accuracy {\bf (AAA)} \citep{caccia2020online}, and the validation set worst-case accuracy {\bf (WC-Acc)} \citep{de2022continual} for all datasets and methods. 
Acc measures how much ``net'' learning occurs after training and
AAA measures the average model performance over all stages of the learning process.
WC-Acc measures how well a model retains past information, penalizing it for ``forgetting and relearning'' a given task after it is learned. We refer readers to \cref{appendix:evalmetrics} for more detail.

\subsection{Memory-Constrained Online Continual Learning}
\label{sec:exp_limited}

Our main result assesses LPR on online continual learning tasks with limited-memory replay buffers ($|\mathcal M| \in \{1000, 2000, 4000\}$).
We compare various experience replay methods with and without LPR
on Split-CIFAR100 (\cref{tab:c100}), Split-TinyImageNet (\cref{tab:tin}), and Online CLEAR (\cref{tab:clear}).

Across almost all replay buffer sizes, methods, and datasets, LPR consistently improves the performance of experience replay methods across all metrics.
For all replay methods, LPR provides the most improvement on Split-CIFAR100, followed by Online CLEAR then Split-TinyImageNet.
We notice that LPR generally yields a higher performance gain on AAA and WC-Acc compared to Acc. This highlights the fact that LPR's preconditioning yields stronger intermediate performance during the optimization process, which results in stronger final performance.
We additionally visualize this behavior by plotting the changes in model performance metrics during optimization for ER and LPR on all datasets and memory sizes in \cref{appendix:trainingmetrics}.

\begin{table*}[tbp]
    \centering
    \resizebox{0.95\textwidth}{!}{%
    \begin{tabularx}{0.975\textwidth}{|>{\centering\arraybackslash}X||>{\centering\arraybackslash}X|>{\centering\arraybackslash}X|>{\centering\arraybackslash}X||>{\centering\arraybackslash}X|>{\centering\arraybackslash}X|>{\centering\arraybackslash}X|}
        \hline
        & \multicolumn{3}{c||}{Memory Size 1000} & \multicolumn{3}{c|}{Memory Size 2000}\\
        \hline
        Method & Acc & AAA & WC-Acc & Acc & AAA & WC-Acc\\
        \hline
        ER & 0.6511 {\scriptsize ± 0.0023} & 0.6433 {\scriptsize ± 0.0056} & 0.5830 {\scriptsize ± 0.0030} & 0.6837 {\scriptsize ± 0.0048} & 0.6620 {\scriptsize ± 0.0048} & 0.6032 {\scriptsize ± 0.0041}\\
        LPR (ER) & \textbf{0.6624} {\scriptsize ± 0.0030} & \textbf{0.6675} {\scriptsize ± 0.0044} & \textbf{0.6406} {\scriptsize ± 0.0036} & \textbf{0.6855} {\scriptsize ± 0.0039} & \textbf{0.6845} {\scriptsize ± 0.0043} & \textbf{0.6552} {\scriptsize ± 0.0053}\\
        \hline
        DER & 0.6655 {\scriptsize ± 0.0125} & 0.6556 {\scriptsize ± 0.0107} & 0.6272 {\scriptsize ± 0.0126} & 0.6980 {\scriptsize ± 0.0109} & 0.6742 {\scriptsize ± 0.0115} & 0.6412 {\scriptsize ± 0.0106}\\
        LPR (DER) & \textbf{0.6804} {\scriptsize ± 0.0176} & \textbf{0.6674} {\scriptsize ± 0.0184} & \textbf{0.6439} {\scriptsize ± 0.0187} & \textbf{0.7147} {\scriptsize ± 0.0175} & \textbf{0.6859} {\scriptsize ± 0.0176} & \textbf{0.6595} {\scriptsize ± 0.0178}\\
        \hline
        EMA & 0.7142 {\scriptsize ± 0.0023} & 0.6508 {\scriptsize ± 0.0037} & 0.5924 {\scriptsize ± 0.0017} & 0.7357 {\scriptsize ± 0.0034} & 0.6672 {\scriptsize ± 0.0034} & 0.6145 {\scriptsize ± 0.0026}\\
        LPR (EMA) & \textbf{0.7406} {\scriptsize ± 0.0023} & \textbf{0.6906} {\scriptsize ± 0.0036} & \textbf{0.6562} {\scriptsize ± 0.0039} & \textbf{0.7558} {\scriptsize ± 0.0013} & \textbf{0.7099} {\scriptsize ± 0.0038} & \textbf{0.6764} {\scriptsize ± 0.0038}\\
        \hline
    \end{tabularx}
    }
    \caption{Online CLEAR results with mean and standard error computed across 5 random seeds.}
    \label{tab:clear}
\end{table*}


\subsection{Memory-Unconstrained Online Continual Learning}
\label{sec:exp_unlimited}

\begin{table}[t!]
    \centering
    \resizebox{0.975\linewidth}{!}{%
    \begin{tabularx}{0.55\textwidth}{|>{\centering\arraybackslash}X|>{\centering\arraybackslash}X|>{\centering\arraybackslash}X|>{\centering\arraybackslash}X|}
        \hline
        \multicolumn{4}{|c|}{Split-CIFAR100}\\
        \hline
        Method & Acc & AAA & WC-Acc \\
        \hline
        ER & 0.3541 {\scriptsize ± 0.0096} & 0.3872 {\scriptsize ± 0.0073} & 0.1391 {\scriptsize ± 0.0028}\\
        LPR (ER) & \textbf{0.4244} {\scriptsize ± 0.0051} & \textbf{0.4644} {\scriptsize ± 0.0040} & \textbf{0.2260} {\scriptsize ± 0.0057}\\
        \hline
        \multicolumn{4}{c}{}\\
        \hline
        \multicolumn{4}{|c|}{Split-TinyImageNet}\\
        \hline
        Method & Acc & AAA & WC-Acc \\
        \hline
        ER & 0.3674 {\scriptsize ± 0.0038} & 0.4070 {\scriptsize ± 0.0073} & 0.2288 {\scriptsize ± 0.0015}\\
        LPR (ER) & \textbf{0.3686} {\scriptsize ± 0.0055} & \textbf{0.4153} {\scriptsize ± 0.0075} & \textbf{0.2437} {\scriptsize ± 0.0025}\\
        \hline
        \multicolumn{4}{c}{}\\
        \hline
        \multicolumn{4}{|c|}{Online CLEAR}\\
        \hline
        Method & Acc & AAA & WC-Acc \\
        \hline
        ER & 0.7176 {\scriptsize ± 0.0062} & 0.6897 {\scriptsize ± 0.0040} & 0.6244 {\scriptsize ± 0.0052}\\
        LPR (ER) & \textbf{0.7411} {\scriptsize ± 0.0040} & \textbf{0.7120} {\scriptsize ± 0.0031} & \textbf{0.6834} {\scriptsize ± 0.0040}\\
        \hline
    \end{tabularx}
    }
    \caption{Online continual learning results with no memory constraints---%
    all prior data are stored in the replay buffer.
    Mean and standard error are computed across 5 random seeds.}
    \label{tab:full}
\end{table}

The previous results demonstrate that LPR improves continual learning performance regardless of the replay buffer size.
To further accentuate this finding,
we test LPR in the limiting case of unlimited replay memory
(i.e. the replay buffer stores all prior training data).
In this setting, we assume that catastrophic forgetting is minimal because the model has access to all prior training data (in expectation) at every point during training.


\cref{tab:full} displays the performance of memory-unconstrained ER with and without LPR on the three previous datasets.
From this table, we can observe many trends.
As expected, the models with unlimited replay memory perform better than their counterparts with limited memory.
This difference suggests that some degree of forgetting occurs with finite replay memory.
Nevertheless, the relative improvement provided by LPR in the memory-unconstrained case largely matches that in the memory-constrained case,
despite the fact that the network has access to all prior data.
The final test accuracy improvement is remarkably large on Split-CIFAR100 dataset, where adding LPR leads to a $7$\% improvement.
Moreover, LPR yields improvements to the AAA and WC-Acc metrics across all three datasets.
These results highlight that, while catastrophic forgetting is a big roadblock to online continual learning, there are non-trivial orthogonal gains to be had from modifying the optimizer.



\subsection{Analysis}

\paragraph{Internal representations and accuracy.}

\begin{figure*}[tbp]
    \centering
    \begin{subfigure}[b]{0.325\linewidth}
        \centering
        \includegraphics[width=\linewidth]{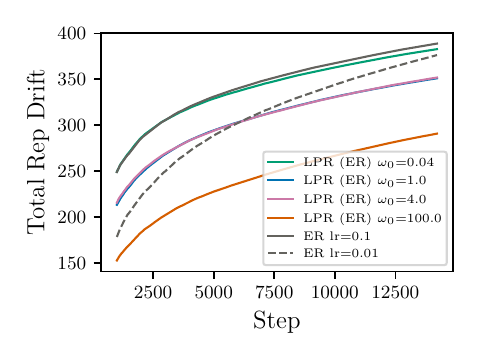}
    \end{subfigure}
    \hspace{0.0025\linewidth}
    \begin{subfigure}[b]{0.325\linewidth}
        \centering
        \includegraphics[width=\linewidth]{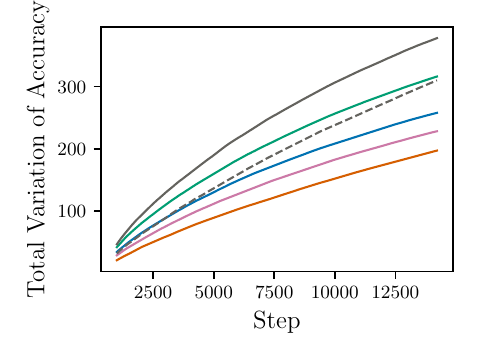}
    \end{subfigure}
    \hspace{0.0025\linewidth}
    \begin{subfigure}[b]{0.325\linewidth}
        \centering
        \includegraphics[width=\linewidth]{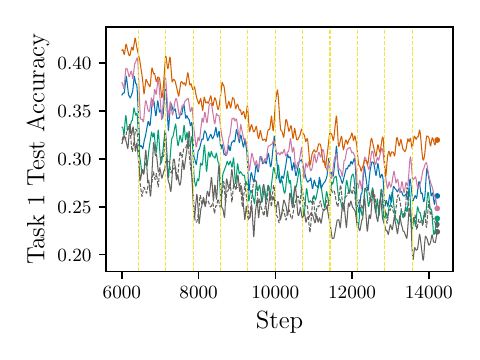}
    \end{subfigure}
    \vspace*{-5mm}
    \caption{Internal representation and accuracy metrics vs entire training iterations for Split-CIFAR100's \emph{task 1 test data}. Results were computed across 5 seeds. All LPR (ER) runs employ a learning rate of 0.1. We mark task boundaries in the right sub-figure using vertical lines. {\bf Left:} dynamics of internal representation changes of task 1 data over the course of training. {\bf Middle:} total variation of task 1 test accuracy. {\bf Right:} task 1 test accuracy. We observe that LPR obtains lower representation drift and lower total variation of accuracy which demonstrate that LPR better preserves predictive stability of past data. This  property is correlated to the overall higher accuracy as shown in the rightmost plot.}
    \label{fig:trdtadacc}
\end{figure*}

LPR's proximal update is designed to ensure that internal representations of past data do not undergo abrupt changes.
To quantify this, we consider a set $\mathcal D$ of test datapoints that correspond to the first Split-CIFAR100 task (i.e. data encountered early in training).
For each test datapoint $i$, we consider its internal representation $Z_i$ as a function of time $j$ (i.e. $Z_i(j)$),
and we measure the stability of these representations using the following metric:
\begin{gather}
    \text{Representation Drift}_\tau = {\textstyle \sum_{j=1}^{\tau-1} \Vert Z_i(j+1) -Z_i(j) \Vert_2},
    \nonumber
\end{gather}
which captures the degree to which each gradient update changes internal representations.
\cref{fig:trdtadacc} (left) demonstrates that networks trained with LPR indeed incur lower average representation drift, suggesting more stable past-data representations throughout training.
Importantly, this slower representational change of LPR cannot simply be replicated by lowering the learning rate for standard gradient descent (dotted line).
Finally, we note that the amount of representation drift is inversely proportional to the $\omega_0$ hyperparameter (see Equation~\ref{eqn:lpr}),
which is expected since LPR converges to standard gradient updates as $\omega_0 \to 0$.
%
%

We find that LPR's representational stability is correlated with predictive stability.
To quantify predictive stability,
we consider accuracy on $\mathcal D$ as a function of time $j$ (i.e. $\mathrm{Acc}_{\mathcal D}(j)$)
and report the total variation of the accuracy over training:
\[
\mathrm{TV}_\tau = {\textstyle \sum_{j=1}^{\tau-1} \vert \mathrm{Acc}_{\mathcal D}(j + 1) - \mathrm{Acc}_{\mathcal D}(j) \vert}, \label{eq:}
\]
which captures the degree to which each gradient update changes predictive performance on past data.
In \cref{fig:trdtadacc} (middle), we observe similar trends to representation drift:
LPR models yield lower total variation of accuracy,
with large $\omega_0$ values corresponding to the least variation.

Crucially, as \cref{fig:trdtadacc} (right) shows, LPR's predictive stability is correlated with higher intermediate and final accuracy.
These results imply that encouraging models to undergo less ``learning'' and ``forgetting'' on sets of data during optimization is an effective strategy for improving model performance.
We however highlight that \cref{fig:trdtadacc} (right) details task 1 test accuracy and not accuracy over all data.
Generally, the higher the value of $\omega_0$ is, the more LPR prioritizes predictive stability over fully adapting the model on the current timestep gradients.
In practice, we find that the optimal $\omega_0$ value is dataset dependent, but typically in the range of $0.25$ to $4$ (refer to Appendix \ref{appendix:experiment}).

\begin{figure}[t!]
    \centering
    \includegraphics[width=\linewidth]{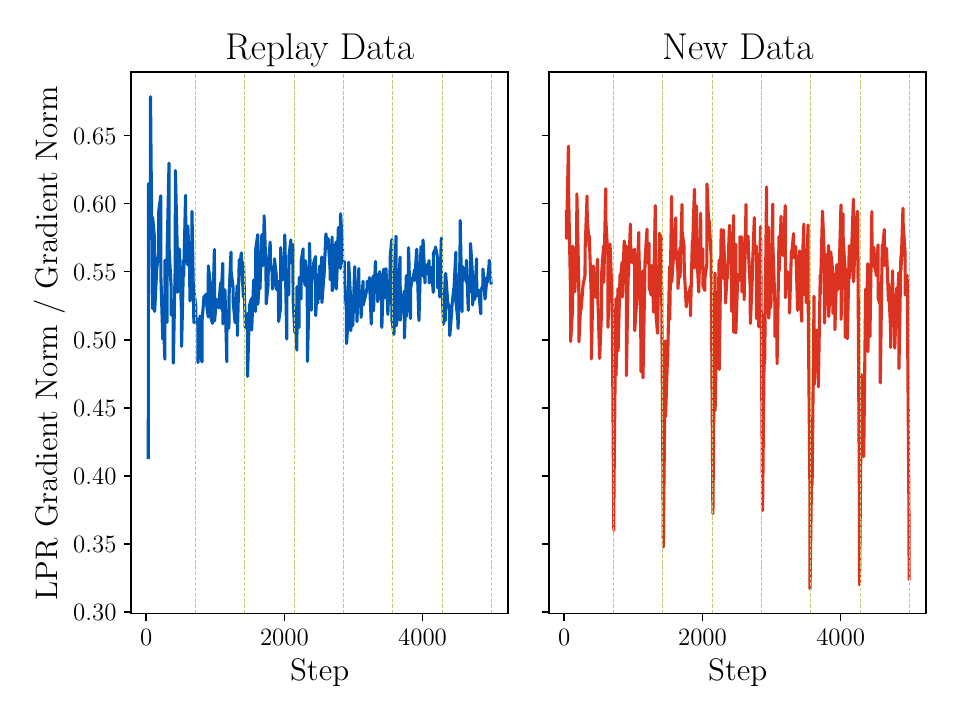}
    \vspace*{-5mm}
    \caption{Ratios between post-preconditioning gradient norms and original gradient norms associated with current and replay data loss vs training iterations for LPR augmented experience replay. The vertical lines denote Split-CIFAR100's task boundaries.}
    \label{fig:gradnorms}
\end{figure}

\paragraph{The effect of LPR preconditioning on gradients.}
Recall from \cref{eqn:lpr} that LPR modifies the standard gradient update with a preconditioner $P_\ell$.
Since this preconditioner is a contraction (i.e. $\Vert P_\ell^{-1} \Vert < 1$)
the LPR parameter updates will always be smaller in magnitude than standard gradient descent updates.
In \cref{fig:gradnorms}, we plot the magnitude ratio between LPR's preconditioned gradients versus standard gradients:
$\Vert P^{-1} \nabla \mathcal L(\theta) \Vert_2 / \Vert \nabla \mathcal L(\theta) \Vert_2$.
We split the gradient contributions of replay data versus new streaming data into the left and right plots respectively.
First, we observe the magnitude ratio generally hovers around $0.55$ for both new and replay data.
This result suggests that LPR has a lower ``effective learning rate'' than that of standard gradient descent, though we suspect that this difference alone does not account for the performance difference.
More strikingly, the relative magnitude of LPR gradients shrinks dramatically whenever the data stream switches to a new task (represented by dotted lines).
For new data, the ``effective learning rate'' decreases to $1/3$
while remaining relatively constant for replay data.
Taken together, these results imply that LPR decreases parameter update magnitude by restricting parameter movement in certain directions whenever gradient updates may yield sudden changes to internal representations and past-data predictions.

\section{Discussion}

In this paper we present Layerwise Proximal Replay (LPR), an online continual learning method that combines experience replay with a proximal point method. We empirically demonstrate that Layerwise Proximal Replay's optimization geometry has non-trivial consequences on the predictive performance of many replay-based online continual learning methods. Importantly, the use of replay buffers and proximal optimizers complement one another. Proximal optimizers perform very poorly without replay (see \cref{appendix:gp_results}), and yet experience replay methods with SGD consistently under-perform their Layerwise Proximal Replay counterparts.

As discussed in \cref{sec:method}, the proximal update proposed in this paper is closely related to projection methods that restrict parameter updates to a subspace orthogonal to past data activations \citep{zeng2019continual,saha2021gradient,deng2021flattening,lin2022trgp,shu2022replay,zhao2023rethinking,saha2023continual}.
In practice, most of these methods do not apply a ``true'' projection to the gradients,
but rather apply ``approximate projections'' that often have functional forms similar to \cref{eqn:woodbury} (see \cref{appendix:prox_baselines}).
We believe that proximal updates, which encompass these projection approximations,
provide a more natural and general space to derive new continual learning methods.

There are several possible avenues for future work.
One is to investigate mechanisms for dynamically adjusting Layerwise Proximal Replay's proximal regularization strength to optimally balance a model's past data prediction stability and plasticity.
Another direction is to make Layerwise Proximal Replay more computationally efficient by updating its preconditioner online rather than periodically recomputing it from scratch.
Finally, extending Layerwise Proximal Replay's proximal formulation to support correlated errors between layer activations, as explored by \citet{duncker2020organizing, kao2021natural} for offline continual learning without experience replay, would be of significant interest.


\newpage

\section*{Acknowledgments}

We acknowledge the support of the Natural Sciences and Engineering Research Council of Canada (NSERC), the Canada CIFAR AI Chairs Program, Inverted AI, MITACS, the Department of Energy through Lawrence Berkeley National Laboratory, and Google. This research was enabled in part by technical support and computational resources provided by the Digital Research Alliance of Canada Compute Canada (alliancecan.ca), the Advanced Research Computing at the University of British Columbia (arc.ubc.ca), and Amazon.

\section*{Impact Statement}
This paper is a step toward effective continual learning of neural networks, which may cause societally impactful deep learning models such as ChatGPT to continuously integrate new data instead of being stuck in the past. Enabling these impactful models to stay up-to-date on the most recent and relevant information can affect how people interact with these models.

\bibliography{example_paper}
\bibliographystyle{icml2024}




\newpage
\appendix
\onecolumn

\section{Layerwise Proximal Replay Derivation}
\label{appendix:preconditioner_proof}

Recall the proximal update from \cref{eq:lagrange},
\begin{align*}
    \theta_{j+1} = \argmin_{\theta} \left( \left\langle \nabla \mathcal L(\theta_j), \theta - \theta_j \right\rangle + \frac{1}{2\eta} \Vert \theta - \theta_j \Vert^2_2 + \sum_{\ell=1}^L  \lambda^{(\ell)} \Vert Z^{(\ell)}_{j} \Theta^{(\ell)} - Z^{(\ell)}_{j} \Theta^{(\ell)}_{j} \Vert_F^2 \right).
\end{align*}
where $\Theta^{(i)} \in \mathbb R^{d^{\ell} \times d^{\ell + 1}}$ is the matrix of parameters at layer $\ell$
such that $\theta = \begin{bmatrix} \mathrm{vec}(\Theta^{(1)}) & \ldots & \mathrm{vec}(\Theta^{(L)}) \end{bmatrix}$,
$\lambda^{(\ell)}$ is the $\ell$-th layer's Lagrange multiplier,
and $Z^{(\ell)}_j \in \mathbb R^{|\mathcal M| \times d^{(\ell)}}$ is the hidden activations for layer $\ell$ at time $j$.

\cref{eq:lagrange} is equivalent to
\begin{align}
    \theta_{j+1} &= \argmin_{\theta}  \left( \sum_{\ell=1}^L  \left\langle \nabla \mathcal L(\Theta_j^{(\ell)}), \Theta^{(\ell)}  - \Theta_j^{(\ell)} \right\rangle
         +  \frac{1}{2\eta} \Vert \Theta^{(\ell)}  - \Theta^{(\ell)}_j \Vert^2_F + \lambda^{(\ell)} \Vert Z^{(\ell)}_{j} \Theta^{(\ell)} - Z^{(\ell)}_{j} \Theta^{(\ell)}_{j} \Vert_F^2  \right),
\end{align}
which entails that the minimization problem w.r.t. $\theta$ can be solved in a layer-wise fashion as follows.
\begin{align}
    \Theta^{(\ell)}_{j+1} &= 
    \argmin_{\Theta^{(\ell)}} 
    \left(  \left\langle \nabla \mathcal L(\Theta_j^{(\ell)}), \Theta^{(\ell)}  - \Theta_j^{(\ell)} \right\rangle
         +  \frac{1}{2\eta} \Vert \Theta^{(\ell)}  - \Theta^{(\ell)}_j \Vert^2_F + \frac{\omega^{\ell}}{2\eta} \Vert Z^{(\ell)}_{j} \Theta^{(\ell)} - Z^{(\ell)}_{j} \Theta^{(\ell)}_{j} \Vert_F^2\right)
\end{align}
where $\omega^\ell = 2\eta\lambda^{(\ell)}$. Using the definition of the Frobenius norm, we have
\begin{align}
    \Theta^{(\ell)}_{j+1} &= 
    \argmin_{\Theta^{(\ell)}} 
    \bigg( \left\langle \nabla \mathcal L(\Theta_j^{(\ell)}), \Theta^{(\ell)}  - \Theta_j^{(\ell)} \right\rangle + \frac{1}{2\eta} \mathrm{tr}\big( \tp{(\Theta^{(\ell)}  - \Theta^{(\ell)}_j)}(\Theta^{(\ell)}  - \Theta^{(\ell)}_j) \big)\\
    &\qquad\qquad\qquad+ \frac{\omega^{\ell}}{2\eta} \mathrm{tr}\big( \tp{(Z^{(\ell)}_{j} \Theta^{(\ell)} - Z^{(\ell)}_{j} \Theta^{(\ell)}_{j})} (Z^{(\ell)}_{j} \Theta^{(\ell)} - Z^{(\ell)}_{j} \Theta^{(\ell)}_{j}) \big) \bigg)\nonumber\\
    &= \argmin_{\Theta^{(\ell)}} 
    \bigg( \left\langle \nabla \mathcal L(\Theta_j^{(\ell)}), \Theta^{(\ell)}  - \Theta_j^{(\ell)} \right\rangle + \frac{1}{2\eta} \mathrm{tr}\big( \tp{(\Theta^{(\ell)}  - \Theta^{(\ell)}_j)} (I + \omega^\ell \tp{Z_j^{(\ell)}} Z_j^{(\ell)}) (\Theta^{(\ell)} - \Theta^{(\ell)}_j) \big) \bigg)\\
    &= \argmin_{\Theta^{(\ell)}} 
    \bigg(  \left\langle \nabla \mathcal{L}(\Theta_j^{(\ell)}), \Theta^{(\ell)}  - \Theta_j^{(\ell)} \right\rangle + \frac{1}{2\eta} \Vert \Theta^{(\ell)}  - \Theta^{(\ell)}_j \Vert^2_{P_\ell} \bigg), \qquad P_\ell := I + \omega^{\ell} Z^{(\ell)^\top}_j Z^{(\ell)}_j \label{eq:layer_obj}.
\end{align}
Setting the derivative of the term inside $\argmin$ to $0$, we have
\begin{gather}
    0 = \nabla \mathcal L(\Theta_j^{(\ell)}) + \frac{1}{\eta} (\omega^{\ell} \tp{Z^{(\ell)}_{j}} Z^{(\ell)}_{j} + I) (\Theta^{(\ell)}  - \Theta^{(\ell)}_j)\\
    \Theta^{(\ell)} = \Theta_j^{(\ell)} - \eta (\omega^{\ell} \tp{Z^{(\ell)}_{j}} Z^{(\ell)}_{j} + I)^{-1} \nabla \mathcal L(\Theta_j^{(\ell)}),
\end{gather}
which concludes the proof.

\section{The Incompatibility of Replay Data and Gradient Projections}
\label{appendix:replay_grad_projections}
In \cref{sec:method}, we argue that gradient projection methods are largely incompatible with replay methods when additional learning on the replay buffer is desired, as is especially the case in online continual learning.
To see why this,
consider the orthogonal projection matrix
$(I - \Phi^{(\ell)} (\Phi^{(\ell)^\top} \Phi^{(\ell)})^{-1} \Phi^{(\ell)^\top})$,
where the matrix
$\Phi^{(\ell)} \in \mathbb R^{d^{(\ell)} \times k}$
defines the $k$-dimensional nullspace of the projection.
For most gradient projection methods, $\Phi^{(\ell)}$ is some compressed representation of past replay activations.

Let $Z^{(\ell)} = \begin{bmatrix} z_1^{(\ell)\top}; \ldots; z_m^{(\ell)\top} \end{bmatrix} \in \mathbb R^{m \times d^{(\ell)}}$ be the activations of $m$ replay datapoints
used to compute the replay loss.
We can decompose these activations as
\[
Z^{(\ell)\top} = \Phi^{(\ell)} A + R,
\]
where $A \in \mathbb R^{k \times m}$ and $R$ is some matrix orthogonal to $\Phi^{(\ell)}$.
If $\Phi^{(\ell)}$ is a good approximation of all replay activations,
then we expect $\Vert R \Vert_F$ to be small.

When computing the replay loss,
note that $Z^{(\ell)}$ interacts with the parameters $\Theta^{(\ell)}$ through the matrix multiplication
$Z^{(\ell)} \Theta^{(\ell)}$.
Therefore, applying standard vector-Jacobian product backpropagation rules,
the replay loss gradient is given as
\[
    \nabla_\Theta^{(\ell)} \mathcal L_\mathrm{replay}
    = Z^{(\ell)\top} V
    = \left( \Phi^{(\ell)} A + R \right) V
\]
for some matrix $V \in \mathbb R^{m \times d^{(\ell + 1)}}$
The projected gradient is thus equal to:
\begin{align*}
    \left( I - \Phi^{(\ell)} (\Phi^{(\ell)^\top} \Phi^{(\ell)})^{-1} \Phi^{(\ell)^\top} \right)
    \nabla_\Theta^{(\ell)} \mathcal L_\mathrm{replay}
    = \left( I - \Phi^{(\ell)} (\Phi^{(\ell)^\top} \Phi^{(\ell)})^{-1} \Phi^{(\ell)^\top} \right) \left( \Phi^{(\ell)} A + R \right) V
    = R V.
\end{align*}
If $\Vert R \Vert_F \approx 0$
(which again will be the case if $\Phi^{(\ell)}$ is a good approximation of all replay activations),
then almost all of the replay gradient will be projected out.

\section{Extended Related Work}
\label{sec:extended_related_work}

There are two papers that employ orthogonal gradient projection for online continual learning: A-GEM \citep{chaudhry2018efficient} and AOP \citep{guo2022adaptive}. Neither of them employs experience replay. AOP is perhaps the closest method to LPR since it employs an incremental algorithm to approximate the layerwise preconditioner $P^\ell = (\tp{Z^\ell}Z^\ell + \alpha I)^{-1}$ with a very low $\alpha$, whose form is identical to LPR's layerwise preconditioner up to a constant factor. However, there are crucial methodological and empirical differences between LPR and AOP. AOP does not use experience replay and their $Z^\ell$ consists of a mini-batch's average activations. In addition, because past data activations in $Z^\ell$ are never updated to the updated model's feed-forward activations of the same data, AOP's $Z^\ell$ becomes stale over time. Last but not least, AOP severely underperforms compared to replay methods in all of our evaluations.

Our work is also related to prior work that highlights the importance of how neural networks are optimized during continual learning with or without memory constraints \citep{ash2020warm, kao2021natural, de2022continual, konishi2023parameter}. \citet{kao2021natural} combines online Laplace approximation and the K-FAC \citep{martens2015optimizing} preconditioner to tackle offline continual learning. A concurrent work from \citet{hess2023two} also proposed the use of orthogonal gradient projection methods with experience replay motivated by the stability gap phenomenon \citep{de2022continual} for both offline and online continual learning. However, they do so without experimental results as of this paper's writing.

\subsection{The Proximal Perspective of OWM and GPM Parameter Updates}
\label{appendix:prox_baselines}
Consider the proximal update
\begin{align}
    \Theta^{(\ell)}_{j+1} 
     &= 
    \argmin_{\Theta^{(\ell)}} 
    \left(  \left\langle \nabla \mathcal L(\Theta_j^{(\ell)}), \Theta^{(\ell)}  - \Theta_j^{(\ell)} \right\rangle
         +  \frac{1}{2\eta} \Vert \Theta^{(\ell)}  - \Theta^{(\ell)}_j \Vert^2_2 + \frac{1}{2\eta} \sum_{m \in M^\ell} \omega_m \Vert m\Theta^{(\ell)} - m\Theta^{(\ell)}_j \Vert_2^2 \right). \label{eq:baseline_obj}
\end{align}

where $\Theta_j^{(\ell)} \in \mathbb{R}^{d^{\ell} \times d^{\ell+1}}$ is the layer $l$ weight matrix at optimization step $j$, $\nabla \mathcal L(\Theta_j^{(\ell)})$ is the objective gradient, $M^\ell \in \mathbb{R}^{n^\ell \times d^{\ell}}$ is the size $n^\ell$ ``representative set'' of $\ell$-th layer's past activations $m \in \mathbb{R}^{1 \times d^{\ell}}$, $\omega_m \in \mathbb{R}_{\geq 0}$ is $m$'s importance weight, and $\eta$ is the learning rate. The representative set $M^\ell$ is allowed to contain pseudo-activation that may not have been observed before. We note \cref{eq:baseline_obj} is a ``weighted version'' of the LPR proximal update where error on different entries $Z^{(l)}$ are penalized to different degrees, dependent on $\omega_m$.

This proximal update, with appropriate choices of $M^\ell$, $\omega_m$, and $\mathcal L(\Theta_j^{(\ell)})$,
recovers OWM-based methods' projected gradient update \citep{zeng2019continual, guo2022adaptive}
\begin{gather}
    \Theta^{(\ell)}_{j+1} = \Theta_{j}^{(\ell)} - \eta (I - \tp{M^\ell} (\alpha I + M^\ell\tp{M^\ell})^{-1} M^\ell) \nabla \mathcal L(\Theta_j^{(\ell)})
\end{gather}
where $\alpha > 0$, and GPM-based methods' scaled projected gradient update \citep{saha2021gradient, deng2021flattening, shu2022replay, saha2023continual}
\begin{gather}
    \Theta^{(\ell)}_{j+1} = \Theta_{j}^{(\ell)} - \eta (I - \tp{M^\ell} \Lambda^\ell M^\ell) \nabla \mathcal L(\Theta_j^{(\ell)})
\end{gather}
where $\Lambda^\ell \in \mathbb{R}^{k \times k}$ is a diagonal matrix whose entries are in $(0, 1]$ and $k < d^{l-1}$.

\begin{proof}
The derivative of the proximal objective in \cref{eq:baseline_obj} is
\begin{gather}
    \nabla \mathcal L(\Theta_j^{(\ell)}) + \frac{1}{\eta} \left(I + \sum_{m \in M^\ell} \omega_m m^{\top}m\right)(\Theta^{(\ell)} - \Theta_j^{(\ell)}).
\end{gather}
Setting this to $0$ and rearranging the terms, we have
\begin{align}
    \Theta^{(\ell)} &= \Theta_j^{(\ell)} - \eta \left( I + \sum_{m \in M^\ell} \omega_m m^{\top}m \right)^{-1} \nabla \mathcal L(\Theta_j^{(\ell)})\\
    &= \Theta_j^{(\ell)} - \eta \left( I + \tp{M^\ell} \Omega M^\ell \right)^{-1} \nabla \mathcal L(\Theta_j^{(\ell)}) \label{eq:proof_lpr}
\end{align}
where $\Omega \in \mathbb{R}^{k \times k}$ is a diagonal matrix with $\omega_k$ on its diagonal. Using Woodbury matrix identity, we then have
\begin{gather}
    \Theta^{(\ell)} = \Theta_j^{(\ell)} - \eta \left( I - \tp{M^\ell} \Lambda^\ell M^\ell \right)^{-1} \nabla \mathcal L(\Theta_j^{(\ell)}) \label{eq:proof_proj_update}\\
    \Lambda^\ell \coloneqq (\Omega^{-1} + M^\ell \tp{M^\ell})^{-1}. \label{eq:proof_proj_lambda}
\end{gather}

If we set the rows of $M^\ell$ to the average embedding of mini-batch for all past mini-batches (each embedding computed right after training on that mini-batch), set $\Omega^{-1} = \alpha I$, and set $\mathcal L(\Theta_j^{(\ell)})$ to the current data batch loss (without replay loss) in Equation \ref{eq:proof_proj_lambda}, and we recover OWM's projected gradient update.

If we set the rows of $M^\ell$ to core gradient space's \citep{saha2021space} orthonormal basis vectors, set diagonal elements of $\Omega^{-1}$ to the soft projection scores between $0$ and $\infty$ associated with each basis vectors, and set $\nabla \mathcal L(\Theta_j^{(\ell)})$ to GPM method specific loss in Equation \ref{eq:proof_proj_lambda}, we recover GPM's projected gradient update since the diagonal entries of $\Lambda^\ell = (\Omega^{-1} + M^\ell \tp{M^\ell})^{-1}$ becomes $\lambda_i = \frac{\omega_i}{\omega_i+1}$. The basis vector importance weight $\omega_i$ can individually be set between $0$ and $\infty$ to cause $\lambda_i$ to be between $[0, 1]$.
\end{proof}




\section{Parameterizing Layerwise Importance Weights}
\label{appendix:omega_param}

We propose a simple method for setting the proximal penalty strength hyperparameters $\omega_0^\ell$ for every layer using two scalar hyperparameters $\omega_0, \beta$. The scalar $\omega_0 \geq 0$ represents the base proximal penalty strength for all network layers. The scalar $\beta \geq 0$ controls the magnitude of a per-layer scaling term $c^\ell$ that is applied to $\omega_0$ based on how many ``effective activation vectors'' can be constructed from a single replay buffer exemplar for matrix multiplication for that layer. We denote this per-layer effective activation vector count by $n_{eff}^\ell$ and note that $n_{eff}^\ell$ is predetermined for each layer.

Concretely, the formula for $\omega_0^\ell$ is
\begin{equation}
    \omega_0^\ell = \frac{\omega_0}{c^\ell}, \quad\quad\quad c^\ell = (n_{eff}^\ell)^{\beta}.
\end{equation}

 We give specific examples of $n_{eff}^\ell$ for linear, 2D convolution, and 2D batch normalization layers, all of which are used in ResNet18. For ResNet18's linear layer of depth $l$, a single replay buffer exemplar translates to a single input embedding of size $d^{l-1}$. Therefore, $n_{eff}^\ell$ is simply $1$. For ResNet18's 2D convolution layer of depth $l$, a single replay buffer exemplar translates to an input feature map of size $C_{\text{in}} \times W_{\text{f}} \times H_{\text{f}}$, where $C_{\text{in}}$ refers to the input channel dimension, $W_{\text{f}} $ and $H_{\text{f}}$ represents the feature map's width and height. To convert 2D convolution on this feature map to matrix multiplication, the input feature map must be converted to a matrix shape $N_{\text{P}} \times (C_{\text{in}} * W_{\text{k}}* H_{\text{k}})$ where $N_{\text{P}}$ denotes the number of patches from the input feature map to be matrix multiplied with the convolution weights, $W_{\text{k}}$ and $H_{\text{k}})$ denotes convolution kernel's width and height.  Therefore, $n_{eff}^\ell$ is equal to $N_{\text{P}}$ specific to the $l$-th layer's convolution operation. Lastly, for ResNet18's 2D batch normalization layer of depth $l$, a single replay buffer exemplar again translates to an input feature map of size $C_{\text{in}} \times W_{\text{f}} \times H_{\text{f}}$. Since 2D batch normalization applies a channel-wise normalization and scaling, it can be seen as $C_{\text{in}}$ independent 1D linear map. For each 1D linear map, the corresponding channel's feature map is independently converted to a vector of size $(W_{\text{f}} * H_{\text{f}})$. Therefore, $n_{eff}^\ell$ is equal to $(W_{\text{f}} * H_{\text{f}})$ for 2D batch normalization.

\section{Additional Experiment Details}
\label{appendix:experiment}

We discuss the datasets we evaluate on in more detail. Split-CIFAR100 dataset is based on \citep{krizhevsky2009learning}, which contains 50,000 training images of size 32x32 that belong to 100 classes. Split-TinyImageNet dataset is based on \cite{le2015tiny}, which contains 100,000 training images of size 64x64 that belong to 200 classes. The CLEAR dataset \citep{lin2021clear} contains 33,000 training images of size 224 x 224 that belong to 11 classes. CLEAR is designed to have a smooth temporal evolution of visual concepts with real-world imagery, which is a scenario more likely to be encountered in real life than the commonly used Permuted MNIST based off \citep{lecun1998gradient} or the Core50 domain-incremental learning benchmark \citep{lomonaco2017core50}.

Hyperparameters for all baselines and LPR augmented baseline runs from the paper were selected once per dataset, based on medium memory size online continual learning experiments (memory size 2000 for Split CIFAR100, 4000 for Split TinyImageNet, and 2000 for Online CLEAR). For each baseline method, we searched across method-specific hyperparameters and selected the configuration that yielded the best final average validation accuracy on a single seed. For LPR augmented baseline methods, we searched across LPR hyperparameters $\omega_0, \beta$ on top of the selected baseline configuration per method in a likewise fashion.

For all baseline methods on all datasets, we searched across their learning rates between $\{0.01, 0.05, 0.1\}$. For DER, we additionally searched across the replay loss weight between $\{0.1, 0.5, 1.\}$ and the logit matching loss weight between $\{0.1, 0.5, 1.\}$. For EMA, we additionally searched across the exponential moving average momentum between $\{0.99, 0.995, 0.999\}$. For LODE, we searched across the old/new class distinction loss weighting $\rho$ between $\{0., 0.05, 0.1\}$. We note that setting $\rho=0.$ recovers ER-ACE \citep{caccia2021new}.

For all LPR runs on all datasets, we searched across $\omega_0$ between $\{0.04, 0.25, 1., 4., 100.\}$ and $\beta$ between $\{1., 2.\}$ on top of the selected baseline configurations. The preconditioner update interval was set to $T=10$ for all experiments, meaning that we updated the preconditioners once every $10$ new data batches.

\cref{tab:main_hyper_lr,tab:main_hyper_omega,tab:main_hyper_beta} contains the selected hyperparameters for replay baselines and its LPR augmentations. In addition to these hyperparameters, we also present replay baseline-specific hyperparameters. DER's replay loss and logit matching loss weights were $1$ and $1$ for Split-CIFAR100, $1$ and $0.5$ for Split-TinyImageNet, and $0.5$ and $1$ for Online CLEAR. EMA's momentum was set to $0.999$ for all datasets. LODE's $\rho$ was set to $0.1$ for Split-CIFAR100 and Split-TinyImageNet.
\begin{table}[ht]
\centering
\begin{tabularx}{0.8\linewidth}{|>{\centering\arraybackslash}X|>{\centering\arraybackslash}X|>{\centering\arraybackslash}X|>{\centering\arraybackslash}X|}
\hline
Method & Split-CIFAR100  & Split-TinyImageNet & Online CLEAR \\ \hline
ER   & 0.1  & 0.01 & 0.01\\
\hline
DER  & 0.05 & 0.01 & 0.1 \\
\hline
EMA  & 0.05 & 0.01 & 0.05\\
\hline
LODE & 0.1  & 0.01 & -   \\
\hline
\end{tabularx}
\caption{Selected learning rates for experience replay baselines.}
\label{tab:main_hyper_lr}
\end{table}

\begin{table}[ht]
\centering
\begin{tabularx}{0.8\linewidth}{|>{\centering\arraybackslash}X|>{\centering\arraybackslash}X|>{\centering\arraybackslash}X|>{\centering\arraybackslash}X|}
\hline
Method & Split-CIFAR100  & Split-TinyImageNet & Online CLEAR \\ \hline
LPR (ER)   & 4 & 0.25 & 1 \\ \hline
LPR (DER)  & 1 & 0.25 & 0.25 \\ \hline
LPR (EMA)  & 0.04 & 0.25 & 0.25\\ \hline
LPR (LODE) & 1 & 0.25 & - \\ \hline
\end{tabularx}
\caption{Selected $\omega_0$ for LPR augmented experience replay baselines.}
\label{tab:main_hyper_omega}
\end{table}

\begin{table}[ht]
\centering
\begin{tabularx}{0.8\linewidth}{|>{\centering\arraybackslash}X|>{\centering\arraybackslash}X|>{\centering\arraybackslash}X|>{\centering\arraybackslash}X|}
\hline
Method & Split-CIFAR100  & Split-TinyImageNet & Online CLEAR \\ \hline
LPR (ER)   & 2 & 2 & 1 \\ \hline
LPR (DER)  & 1 & 2 & 1 \\ \hline
LPR (EMA)  & 1 & 2 & 1 \\ \hline
LPR (LODE) & 2 & 2 & - \\ \hline
\end{tabularx}
\caption{Selected $\beta$ for LPR augmented experience replay baselines.}
\label{tab:main_hyper_beta}
\end{table}

\section{Evaluation Metric Definitions}
\label{appendix:evalmetrics}

We define final accuracy (Acc), average anytime accuracy (AAA) \citep{caccia2020online}, and worst-case accuracy (WC-Acc) metrics \citep{de2022continual}. Let $\tau$ be the current data batch index, $\mathcal{D}_{i}$ be task $i$'s evaluation dataset, and $\mathrm{Acc}_{\mathcal{D}_i}(j)$ be the accuracy of the model trained up to the $j$-th data batch on $\mathcal{D}_i$. In addition, let $\tau_i$ be the index of task $i$'s last data batch in the data stream and $k_\tau$ be the number of tasks the model has been trained on upon training on the $\tau$-th data batch. We then have the following for a model trained up to the $\tau$-th data batch.
\begin{gather}
    \text{Acc} = {\textstyle \frac{1}{k_\tau} \sum_{i=1}^{k_\tau} \mathrm{Acc}_{\mathcal{D}_i}(\tau)} \nonumber,\\
    \text{AAA} = {\textstyle \frac{1}{\tau} \sum_{j=1}^{\tau} \frac{1}{k_j} \sum_{i=1}^{k_j} \mathrm{Acc}_{\mathcal{D}_i}(j)} \nonumber,\\
    \text{WC-Acc} = {\textstyle \frac{1}{k_\tau} [\mathrm{Acc}_{{\mathcal{D}_{k_\tau}}}\!(\tau) + \sum_{i=1}^{k_\tau\!-\!1}\!\min_{\tau_{i} < j \leq \tau}\!\mathrm{Acc}_{\mathcal{D}_i}(j)] }. \nonumber
\end{gather}

\section{LPR Sensitivity Analysis}
\label{appendix:sensitivity}

\begin{table}[h!]
    \centering
    \resizebox{0.95\linewidth}{!}{%
    \begin{tabularx}{0.55\linewidth}{|>{\centering\arraybackslash}X|>{\centering\arraybackslash}X|>{\centering\arraybackslash}X|>{\centering\arraybackslash}X|}
        \hline
        \multicolumn{4}{|c|}{Split-CIFAR100 Test Accuracy}\\
        \hline
         & $T = 5$ & $T = 10$ & $T = 50$ \\
        \hline
        $p = 0.05$ & \underline{0.3307} {\scriptsize ± 0.0038} & 0.3295 {\scriptsize ± 0.0038} & 0.3268 {\scriptsize ± 0.0044}\\
        $p = 0.20$ & \textbf{0.3324} {\scriptsize ± 0.0033} & \underline{\textbf{0.3355}} {\scriptsize ± 0.0041} & 0.3310 {\scriptsize ± 0.0062}\\
        $p = 1.00$ & 0.3294 {\scriptsize ± 0.0039} & \underline{0.3334} {\scriptsize ± 0.0056} & \textbf{0.3332} {\scriptsize ± 0.0043}\\
        \hline
        ER & \multicolumn{3}{c|}{0.2937 {\scriptsize ± 0.0054}}\\
        \hline

        \multicolumn{4}{c}{}\\
        \hline
        \multicolumn{4}{|c|}{Split-TinyImageNet Test Accuracy}\\
        \hline
         & $T = 5$ & $T = 10$ & $T = 50$ \\
        \hline
        $p = 0.05$ & \underline{0.2306} {\scriptsize ± 0.0025} & 0.2300 {\scriptsize ± 0.0024} & \textbf{0.2277} {\scriptsize ± 0.0018}\\
        $p = 0.20$ & \underline{0.2321} {\scriptsize ± 0.0024} & 0.2306 {\scriptsize ± 0.0016} & 0.2259 {\scriptsize ± 0.0016}\\
        $p = 1.00$ & \underline{\textbf{0.2324}} {\scriptsize ± 0.0019} & \textbf{0.2312} {\scriptsize ± 0.0023} & 0.2263 {\scriptsize ± 0.0027}\\
        \hline
        ER & \multicolumn{3}{c|}{0.2192 {\scriptsize ± 0.0021}}\\
        \hline
        
        \multicolumn{4}{c}{}\\
        \hline
        \multicolumn{4}{|c|}{Online CLEAR Test Accuracy}\\
        \hline
         & $T = 5$ & $T = 10$ & $T = 50$ \\
        \hline
        $p = 0.05$ & \underline{0.6871} {\scriptsize ± 0.0028} & 0.6800 {\scriptsize ± 0.0070} & 0.6761 {\scriptsize ± 0.0036}\\
        $p = 0.20$ & 0.6828 {\scriptsize ± 0.0023} & 0.6777 {\scriptsize ± 0.0043} & \underline{0.6831} {\scriptsize ± 0.0028}\\
        $p = 1.00$ & \underline{\textbf{0.6914}} {\scriptsize ± 0.0047} & \textbf{0.6855} {\scriptsize ± 0.0039} & \textbf{0.6873} {\scriptsize ± 0.0027}\\
        \hline
        ER & \multicolumn{3}{c|}{0.6837 {\scriptsize ± 0.0048}}\\
        \hline
    \end{tabularx}

    \hspace{0.5cm}

    \begin{tabularx}{0.55\linewidth}{|>{\centering\arraybackslash}X|>{\centering\arraybackslash}X|>{\centering\arraybackslash}X|>{\centering\arraybackslash}X|}
        \hline
        \multicolumn{4}{|c|}{Split-CIFAR100 Total Wallclock Time}\\
        \hline
         & $T = 5$ & $T = 10$ & $T = 50$ \\
        \hline
        $p = 0.05$ & 19 & 13 & 9\\
        $p = 0.20$ & 25 & 16 & 9\\
        $p = 1.00$ & 45 & 28 & 11\\
        \hline
        ER & \multicolumn{3}{c|}{7}\\
        \hline

        \multicolumn{4}{c}{}\\
        \hline
        \multicolumn{4}{|c|}{Split-TinyImageNet Total Wallclock Time}\\
        \hline
         & $T = 5$ & $T = 10$ & $T = 50$ \\
        \hline
        $p = 0.05$ & 75 & 59 & 46\\
        $p = 0.20$ & 119 & 81 & 50\\
        $p = 1.00$ & 278 & 184 & 67\\
        \hline
        ER & \multicolumn{3}{c|}{39}\\
        \hline
        
        \multicolumn{4}{c}{}\\
        \hline
        \multicolumn{4}{|c|}{Online CLEAR Total Wallclock Time}\\
        \hline
        & $T = 5$ & $T = 10$ & $T = 50$ \\
        \hline
        $p = 0.05$ & 186 & 162 & 142\\
        $p = 0.20$ & 265 & 203 & 151\\
        $p = 1.00$ & 482 & 431 & 176\\
        \hline
        ER & \multicolumn{3}{c|}{135}\\
        \hline
    \end{tabularx}
    }
    
    \caption{\textbf{(Left)} Test accuracy comparison LPR (ER) runs with different values of preconditioner update interval $T$ and replay buffer subsampling proportion $p$. The best LPR (ER) runs for a fixed value of $T$ is bolded and the best LPR (ER) runs for a fixed value of $p$ is underlined. ER test accuracy is listed at the bottom of each table for reference. \textbf{(Right)} Average wallclock time comparison for the same set of LPR (ER) runs, measured in \textit{minutes}.}
    \label{tab:ablation}
\end{table}

In this section, we investigate the effect that different values of preconditioner update interval and replay buffer subsampling for preconditioner construction have on LPR's test accuracy and wallclock time. Let $T$ be the number of data batches we observe before recomputing $P_\ell$ from the replay buffer and let $p$ be the proportion of the replay buffer data that we are using to compute $P_\ell$ (ex. $p=0.05$ means we are sampling 5\% of the replay buffer then computing $P_\ell$ from those examples).

We report the test set accuracy and wallclock time for different LPR (ER) runs with replay buffer size 2000 in \cref{tab:ablation}. The results are averaged across 10 random seeds for Split-CIFAR100, Split-TinyImageNet, and averaged across 5 random seeds for Online CLEAR. 

By and large, the performance differences between LPR runs that have different values of $T$ and $p$ are not drastic, which demonstrates that LPR's performance is not very sensitive to the choice of these hyperparameters. We observe that updating the preconditioner more frequently tends to result in better scores, evident from the fact that the $T=5$ column is most frequently underlined across datasets. This is expected since the hidden activations that the proximal regularizer is computed on are more ``up to date'' w.r.t. the current network parameters. In addition, we observe that updating the preconditioner with more replay buffer data tends to result in better scores, evident from the fact that the $p=1.00$ row is most frequently bolded across datasets. This is also expected since the proximal regularizer is applied to a higher number of past datapoint hidden activations. All LPR configurations outperform ER on Split-CIFAR100 and Split-TinyImageNet datasets in a statistically significant manner.

As for computation, lower values of $T$ and higher values of $p$ significantly increase LPR's wallclock time. We note that even at the most computationally minimal setting ($T=50, p=0.05$) where ER and LPR's runtimes are nearly identical (with $\leq 7$ minutes of runtime overhead), LPR significantly outperforms ER on Split-CIFAR100 and Split-TinyImageNet. While the main results of this paper were presented with $T=10$ and $p=1.00$, it would be of interest to investigate how to best select LPR's hyperparameters s.t. the predictive performance is maximized but the runtime is minimized.

\section{Performance Metrics During Optimization}
\label{appendix:trainingmetrics}

\begin{figure*}[h!]
    \centering
    \begin{subfigure}[b]{0.495\linewidth}
        \centering
        \includegraphics[width=\linewidth]{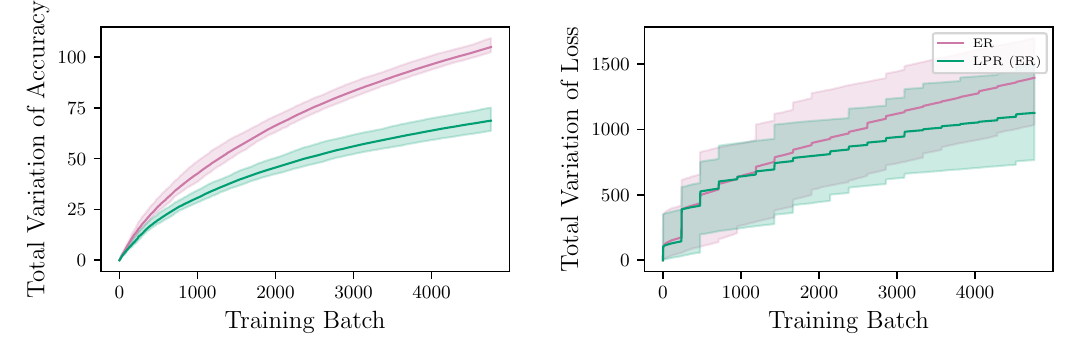}
    \end{subfigure}
    \hfill
    \begin{subfigure}[b]{0.495\linewidth}
        \centering
        \includegraphics[width=\linewidth]{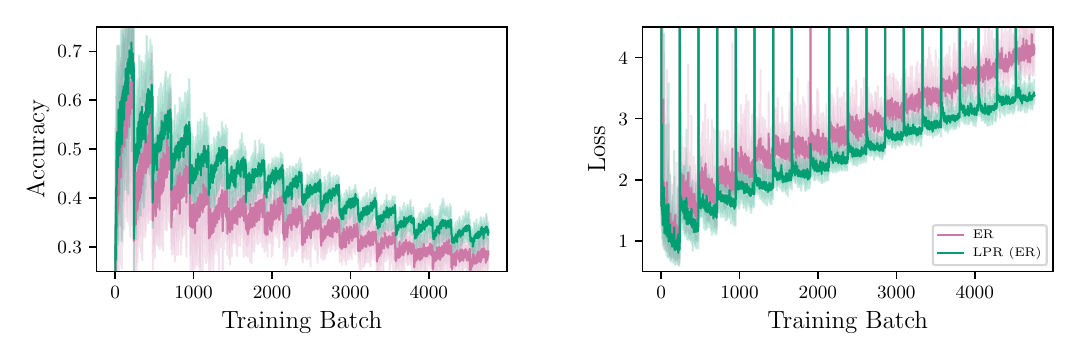}
    \end{subfigure}
    \vspace*{-3mm}
    \caption{Holdout set optimization metrics for Split-CIFAR100 experiments with memory size 2000. Shading denotes the minimum and maximum values observed across 5 seeds.}
    \label{fig:trainplots_c100}
\end{figure*}

\begin{figure*}[h!]
    \centering
    \begin{subfigure}[b]{0.495\linewidth}
        \centering
        \includegraphics[width=\linewidth]{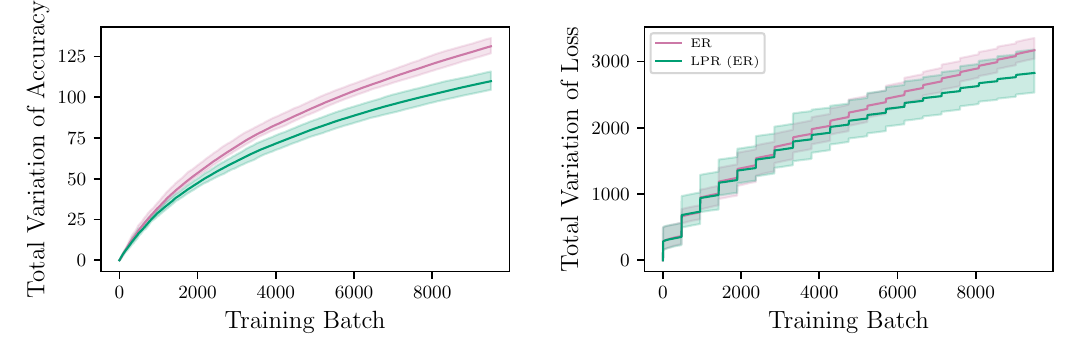}
    \end{subfigure}
    \hfill
    \begin{subfigure}[b]{0.495\linewidth}
        \centering
        \includegraphics[width=\linewidth]{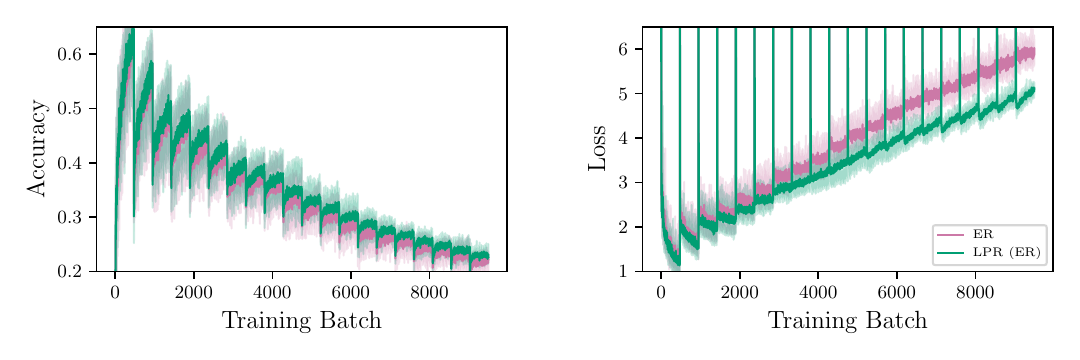}
    \end{subfigure}
    \vspace*{-3mm}
    \caption{Holdout set optimization metrics for Split-TinyImageNet experiments with memory size 4000. Shading denotes the minimum and maximum values observed across 5 seeds.}
    \label{fig:trainplots_tin}
\end{figure*}

\begin{figure*}[h!]
    \centering
    \begin{subfigure}[b]{0.495\linewidth}
        \centering
        \includegraphics[width=\linewidth]{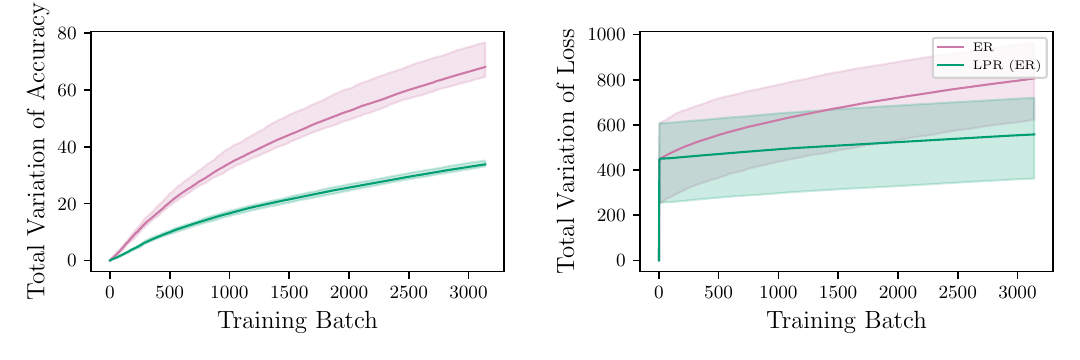}
    \end{subfigure}
    \hfill
    \begin{subfigure}[b]{0.495\linewidth}
        \centering
        \includegraphics[width=\linewidth]{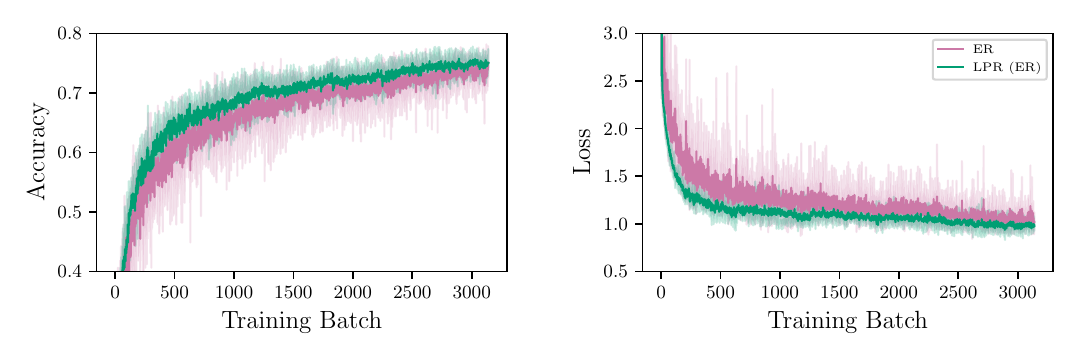}
    \end{subfigure}
    \vspace*{-3mm}
    \caption{Holdout set optimization metrics for Online CLEAR experiments with memory size 2000. Shading denotes the minimum and maximum values observed across 5 seeds.}
    \label{fig:trainplots_clear}
\end{figure*}

\begin{figure*}[h!]
    \centering
    \begin{subfigure}[b]{0.48\linewidth}
        \centering
        \includegraphics[width=\linewidth]{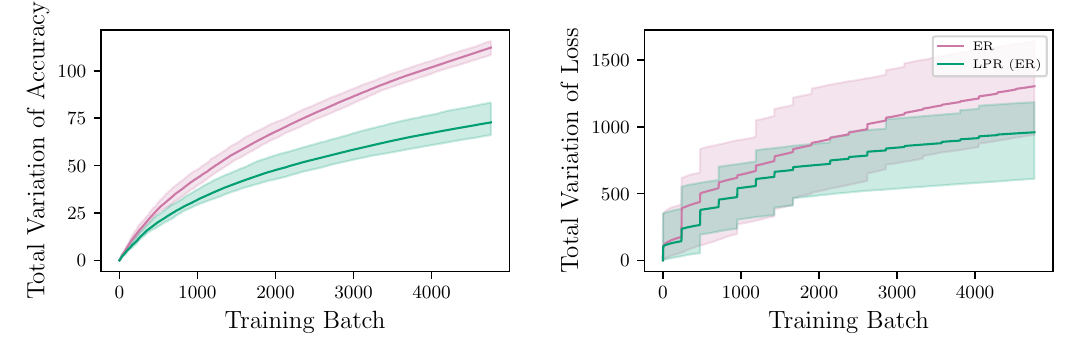}
    \end{subfigure}
    \hfill
    \begin{subfigure}[b]{0.48\linewidth}
        \centering
        \includegraphics[width=\linewidth]{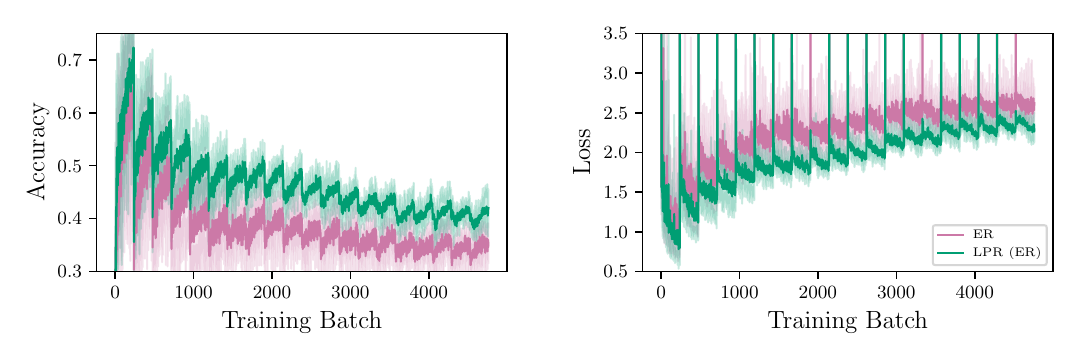}
    \end{subfigure}
    \vspace*{-3mm}
    \caption{Validation set optimization metrics for memory-unconstrained Split-CIFAR100 experiments. Shading denotes the minimum and maximum values observed across 5 seeds.}
    \label{fig:trainplotsfull_c100}
\end{figure*}

\begin{figure*}[h!]
    \centering
    \begin{subfigure}[b]{0.48\linewidth}
        \centering
        \includegraphics[width=\linewidth]{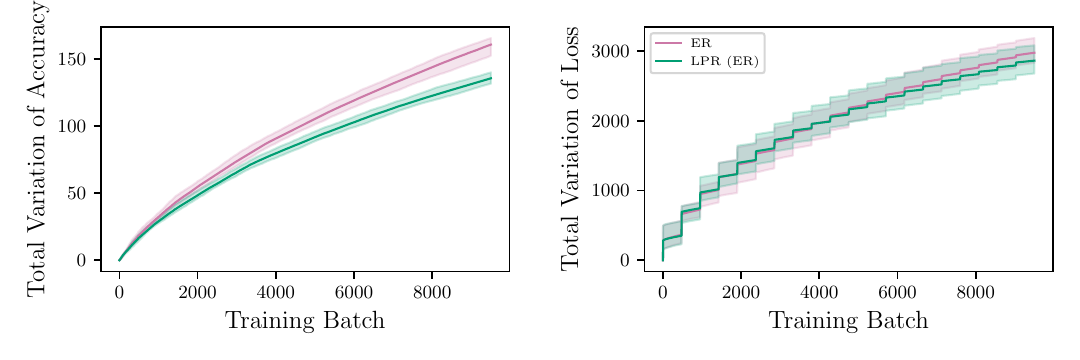}
    \end{subfigure}
    \hfill
    \begin{subfigure}[b]{0.48\linewidth}
        \centering
        \includegraphics[width=\linewidth]{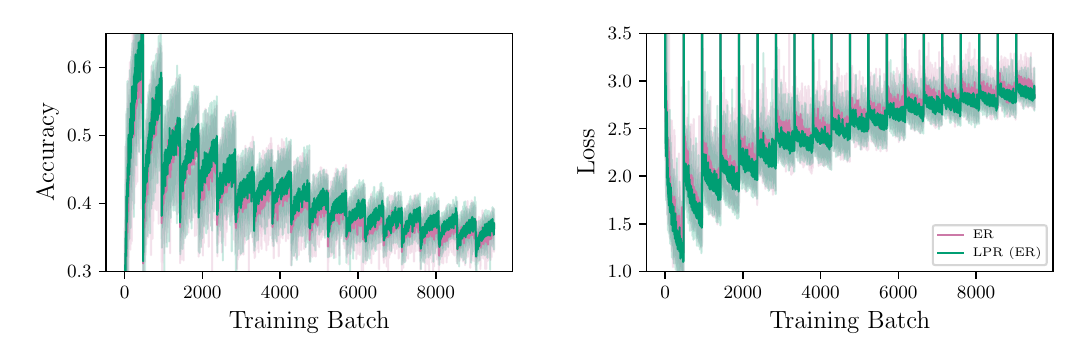}
    \end{subfigure}
    \vspace*{-3mm}
    \caption{Validation set optimization metrics for memory-unconstrained Split-TinyImageNet experiments. Shading denotes the minimum and maximum values observed across 5 seeds.}
    \label{fig:trainplotsfull_tin}
\end{figure*}

\begin{figure*}[h!]
    \centering
    \begin{subfigure}[b]{0.48\linewidth}
        \centering
        \includegraphics[width=\linewidth]{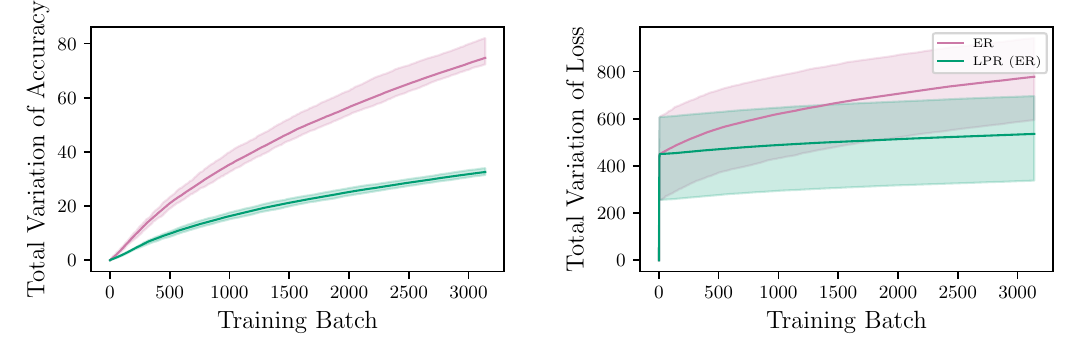}
    \end{subfigure}
    \hfill
    \begin{subfigure}[b]{0.48\linewidth}
        \centering
        \includegraphics[width=\linewidth]{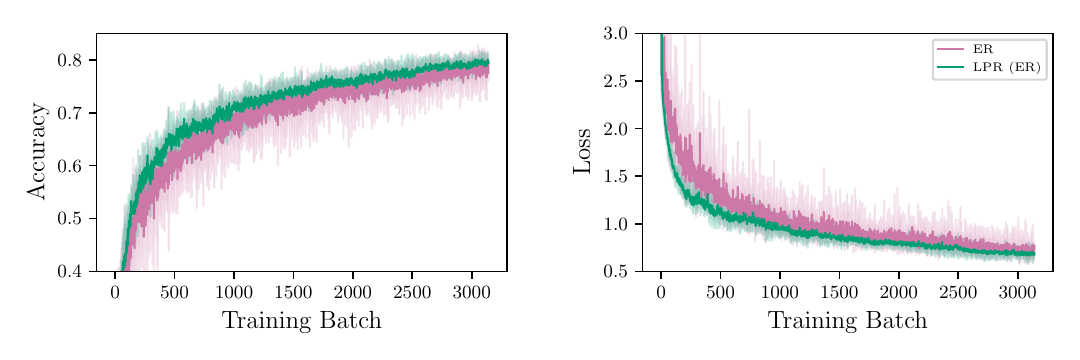}
    \end{subfigure}
    \vspace*{-3mm}
    \caption{Validation set optimization metrics for memory-unconstrained Online CLEAR experiments. Shading denotes the minimum and maximum values observed across 5 seeds.}
    \label{fig:trainplotsfull_clear}
\end{figure*}

To assess LPR's effect on predictive stability and optimization efficiency, we plot the average validation set accuracy, loss, and their total variations for experience replay and LPR-augmented experience replay during optimization.
\cref{fig:trainplots_c100}, \ref{fig:trainplots_tin}, \ref{fig:trainplots_clear} illustrate the results for the memory-constrained experiments and \cref{fig:trainplotsfull_c100}, \ref{fig:trainplotsfull_tin}, \ref{fig:trainplotsfull_clear}
illustrate the results for the memory-unconstrained experiments.
Unlike in \cref{fig:trdtadacc}, the metrics here are averaged across the validation sets from all previously observed tasks at each training timestep.
Therefore, the number of validation set datapoints used to compute these metrics increases whenever new tasks are encountered.
For class-incremental benchmarks, the number of classes the model must predict also increases over time.

The two leftmost columns of all figures show that LPR results in lower total variation of accuracy and loss than ER on all benchmarks throughout optimization.
This demonstrates that LPR’s parameter updates, designed to reduce internal representation drift, enhance neural networks’ predictive stability when learning from non-stationary data streams.
In contrast, the two rightmost columns of all figures show that LPR achieves higher accuracy and lower loss than ER on all benchmarks throughout optimization.
This suggests that LPR’s stabilizing parameter updates improve overall optimization efficiency.
Lastly, we notice that the gap between ER and LPR's validation set loss is wider in the memory-constrained settings.
This suggests that while LPR's optimization benefits are present regardless of replay buffer size, its stabilizing parameter updates may also alleviate catastrophic forgetting to some degree.

\section{Gradient Projection Results on Online Continual Learning}
\label{appendix:gp_results}

\begin{table}[h!]
    \centering
    \resizebox{0.5\linewidth}{!}{%
    \begin{tabularx}{0.55\linewidth}{|>{\centering\arraybackslash}X|>{\centering\arraybackslash}X|>{\centering\arraybackslash}X|>{\centering\arraybackslash}X|}
        \hline
        \multicolumn{4}{|c|}{Split-CIFAR100}\\
        \hline
        Method & Acc & AAA & WC-Acc \\
        \hline
        AOP & 0.0472 {\scriptsize ± 0.0010} & 0.1293 {\scriptsize ± 0.0029} & 0.0411 {\scriptsize ± 0.0022}\\
        AGEM & 0.0327 {\scriptsize ± 0.0024} & 0.1004 {\scriptsize ± 0.0027} & 0.0317 {\scriptsize ± 0.0023}\\
        LPR (ER) & \textbf{0.4244} {\scriptsize ± 0.0051} & \textbf{0.4644} {\scriptsize ± 0.0040} & \textbf{0.2260} {\scriptsize ± 0.0057}\\

        \hline

        \multicolumn{4}{c}{}\\
        \hline
        \multicolumn{4}{|c|}{Split-TinyImageNet}\\
        \hline
        Method & Acc & AAA & WC-Acc \\
        \hline
        AOP & 0.0363 {\scriptsize ± 0.0017} & 0.0929 {\scriptsize ± 0.0018} & 0.0298 {\scriptsize ± 0.0015}\\
        AGEM & 0.0295 {\scriptsize ± 0.002} &0.0683 {\scriptsize ± 0.0033} & 0.025{\scriptsize ± 0.0012}\\
        LPR (ER) & \textbf{0.3686} {\scriptsize ± 0.0055} & \textbf{0.4153} {\scriptsize ± 0.0075} & \textbf{0.2437} {\scriptsize ± 0.0025}\\
        \hline
        \multicolumn{4}{c}{}\\
        \hline
        \multicolumn{4}{|c|}{Online CLEAR}\\
        \hline
        Method & Acc & AAA & WC-Acc \\
        \hline
        AOP & 0.5323 {\scriptsize ± 0.0056} & 0.5596 {\scriptsize ± 0.0055} & 0.5050 {\scriptsize ± 0.0036}\\
        AGEM & 0.5731 {\scriptsize ± 0.0050} & 0.5272 {\scriptsize ± 0.0025} & 0.4748 {\scriptsize ± 0.0015}\\
        LPR (ER) & \textbf{0.7411} {\scriptsize ± 0.0040} & \textbf{0.7120} {\scriptsize ± 0.0031} & \textbf{0.6834} {\scriptsize ± 0.0040}\\
        \hline
    \end{tabularx}
    }
    \caption{Online continual learning results with mean and standard error computed across 5 random seeds.}
    \label{tab:proj}
\end{table}

\begin{table}[ht]
\centering
\begin{tabularx}{0.8\linewidth}{|>{\centering\arraybackslash}X|>{\centering\arraybackslash}X|>{\centering\arraybackslash}X|>{\centering\arraybackslash}X|}
\hline
Method & Split-CIFAR100  & Split-TinyImageNet & Online CLEAR \\ \hline
AGEM & 0.05 & 0.10 & 0.01\\ \hline
AOP & 0.05 & 0.01 & 0.05 \\ \hline
\end{tabularx}
\caption{Optimal learning rates for gradient projection baselines, AGEM and AOP.}
\label{tab:proj_hyper}
\end{table}

We evaluate two gradient projection-based methods, AGEM \citep{chaudhry2018efficient} and AOP \citep{guo2022adaptive}, on the online versions of the Split-CIFAR100 (memory size 2000), Split-TinyImageNet (memory size 4000), and Online CLEAR (memory size 2000) datasets and compare them to our proposed LPR method. The same network architecture is utilized for AGEM and AOP as for LPR, with the optimal learning rate, selected through our hyperparameter search, reported in Table~\ref{tab:proj_hyper}. We report the evaluation results over 5 seeds in Table~\ref{tab:proj}. We observe a huge performance gap between the two projection methods and our LPR method on online class-incremental learning, which we attribute to the fact that the gradient projection methods underfit the data stream. For online domain-incremental learning, LPR still obtains a big performance gain compared to AGEM and AOP.
\end{document}